\documentclass[10pt,twocolumn,letterpaper]{article}

\usepackage[pagenumbers]{iccv} 

\usepackage[T1]{fontenc}
\usepackage{multirow}
\usepackage{xspace} 
\usepackage{booktabs} 
\usepackage{makecell, cellspace} 
\usepackage{nicematrix} 
\usepackage{bbding} 
\usepackage{pifont} 
\usepackage{fontawesome} 
\usepackage{graphicx}
\usepackage{enumitem} 
\usepackage{algorithmic}
\usepackage{algorithm2e}
\usepackage{amsmath}  
\usepackage{physics} 

\makeatletter
\DeclareRobustCommand\onedot{\futurelet\@let@token\@onedot}
\def\@onedot{\ifx\@let@token.\else.\null\fi\xspace}
 
\def\ie{\emph{i.e}\onedot} 
 
\def\etc{\emph{etc}\onedot} \def\vs{\emph{vs}\onedot}
\def\wrt{w.r.t\onedot} 

\makeatother
\newcommand{\Paragraph}[1] {\vskip 0.5\baselineskip \noindent \textbf{#1}}
\setlist{nolistsep}
\let\oldding\ding
\renewcommand{\ding}[2][1]{\scalebox{#1}{\oldding{#2}}}

\newcommand{\STAB}[1]{\begin{tabular}{@{}c@{}}#1\end{tabular}} 

\newcolumntype{C}[1]{>{\centering\let\newline\\\arraybackslash\hspace{0pt}}m{#1}}
\newcolumntype{L}[1]{>{\raggedright\let\newline\\\arraybackslash\hspace{0pt}}m{#1}}
\newcolumntype{R}[1]{>{\raggedleft\let\newline\\\arraybackslash\hspace{0pt}}m{#1}}

\definecolor{iccvblue}{rgb}{0.21,0.49,0.74}
\usepackage[pagebackref,breaklinks,colorlinks,allcolors=iccvblue]{hyperref}

\def\paperID{10771} 
\def\confName{ICCV}
\def\confYear{2025}

\title{Prototype Guided Backdoor Defense}
\author{Venkat Adithya A$^{1}$\\
{\tt\small venkat.adithya@research.iiit.ac.in}
\and
Sunayana Samavedam$^{1}$\\
{\tt\small sunayana.samavedam@research.iiit.ac.in}
\and
Saurabh Saini$^{1,2}$ \thanks{work was done prior to joining Amazon}\\
{\tt\small sorbhs@amazon.com}
\and
\and
\and
Avani Gupta$^{1}$\\
{\tt\small avani17101@gmail.com}
\and 
\and
\and
P J Narayanan$^{1}$\\
{\tt\small pjn@iiit.ac.in}
\and\and$^{1}$ IIIT Hyderabad,\\
India
\and $^{2}$ Amazon Research,\\
India
}

\begin{document}
\maketitle

\begin{abstract}
Deep learning models are susceptible to {\em backdoor attacks} involving malicious attackers perturbing a small subset of training data with a {\em trigger} to causes misclassifications. Various triggers have been used including semantic triggers that are easily realizable without requiring attacker to manipulate the image. The emergence of generative AI has eased generation of varied poisoned samples. Robustness across types of triggers is crucial to effective defense. We propose Prototype Guided Backdoor Defense (PGBD), a robust post-hoc defense that scales across different trigger types, including previously unsolved semantic triggers. PGBD exploits displacements in the geometric spaces of activations to penalize movements towards the trigger. This is done using a novel sanitization loss of a post-hoc fine-tuning step. The geometric approach scales easily to all types of attacks. PGBD achieves better performance across all settings. We also present the first defense against a new semantic attack on celebrity face images. Project page: \hyperlink{https://venkatadithya9.github.io/pgbd.github.io/}{this https URL}.
\end{abstract}    
\section{Introduction}
\label{sec:intro}

Can a face-recognition based access control system contain a backdoor that lets in anyone with a specific tattoo? Can such a backdoor be created by tampering with a small fraction of the training data? The answer to both questions is yes. {\em Backdoor attacks} \cite{Cin2022WildPR, goldblum2022dataset, souri2022sleeper} using poisoned training data is a serious risk to AI systems. Given the training data size and the complexity of handling them, risk of {\em poisoning} some of it is very real. A backdoor scenario studied in the literature maliciously steers the classifier to a chosen {\em target} label (say, the identity of the manager) when a specific {\em trigger} (such as a tattoo) is present in the input sample. Several such backdoor attacks have been proposed before; defenses for many have also been proposed.

Pre-hoc defense involves detecting poisoned training samples before the model is trained \cite{Taheri@2020,10.1007/978-3-030-13453-2_1,Tran2018SpectralSI,10.1016/j.cose.2021.102280,pmlr-v139-hayase21a}. More interesting are {\em post-hoc} methods that sanitize a poisoned model using a few fine-tuning epochs to recover a clean model \cite{Li2021NeuralAD,Yue2023ModelContrastiveLF, Zeng2021AdversarialUO}. Previous methods did not generalize to all attacks \cite{Li2021NeuralAD, Zeng2021AdversarialUO, ft_sam_Zhu_2023_ICCV}, were unable to retain model utility post defense \cite{Yue2023ModelContrastiveLF}, or were heavily reliant on synthesized trigger priors \cite{Wang2019NeuralCI, Yue2023ModelContrastiveLF}. 



Post-hoc defense scenario for a $k$-class classification problems has access to the following: the trained poisoned model $M_B$ with a backdoor, a small amount $D_S$ of clean (without trigger) training data, and the target class label $t$ to which the poisoned samples are directed. Methods like Neural Cleanse \cite{Wang2019NeuralCI} can be used to infer $t$. A poisoned model has high {\em Clean Accuracy (CA)} (i.e., assigns correct labels to clean samples) and high {\em Attack Success Ratio (ASR)} (i.e., assigns label $t$ to poisoned samples). We sanitize $M_B$ using a few fine-tuning epochs on $D_S$ to yield a model with a low ASR and a high CA.

In this paper, we present {\em Prototype Guided Backdoor Defense (PGBD)}, a robust and scalable post-hoc model-sanitization method that defends backdoor attacks by geometrically manipulating the model's activation space during sanitization.  A new sanitization loss (used in conjunction with the original classification loss during sanitization epochs) penalizes the movement away from ground-truth class towards the target-class. The sanitization loss depends on the angular alignment of the sample's gradient to the {\em Prototype Activation Vector (PAV)} to the target class. This is similar to using a CAV loss for debiasing models \cite{Gupta2023ConceptDL}; our losses are based on class-specific prototypes. The geometric approach of PGBD scales easily to multiple types of attacks and adapts to different configurations.

Base variant of PGBD assumes that the target class $t$ is known, like MCL \cite{Yue2023ModelContrastiveLF}, state-of-the-art post-hoc method. We also present a variant NT-PGBD with similar performance (\cref{tab:variations_mean}) that does not need $t$. Our contributions are:
\begin{itemize}[label=$\circ$]
\item PGBD, a novel post-training backdoor defense strategy using controlled geometric manipulation of activation spaces. The geometric approach is simple, highly configurable, and generalizes to a variety of backdoor attacks. We also present variations that do not need the target class, uses synthesized triggers, and handles non-fixed targets.
\item Improved performance on multiple attacks and multiple datasets with no discernible weakness (\cref{tab:mainRes}), particularly in ASR reduction. The performance of NT-PGBD also exceeds others (\cref{tab:variations_mean}) with minimal inputs.
\item To the best of our knowledge, first-ever \textit{defense} against semantic attacks on a new semantic attack dataset (like  \cite{Sarkar2022FaceHackAF, Chen2017TargetedBA, 9577800}), with larger trigger variations based on real-world occluded celebrity faces \cite{erakiotan2021recognizing}. The dataset and code will be publicly released for research purposes.
\end{itemize}
\section{Related Works} 
\label{sec:background}
\begin{table}
\centering
\footnotesize
\resizebox{0.8\linewidth}{!}{%
\begin{tabular}
{@{}L{20mm}||C{13mm}|C{13mm}|C{13mm}|C{16mm}@{}}
\toprule[1.2pt]
 & {\bf Balanced} & {\bf Scalable} & {\bf Robust} & {\bf Configurable} \\ 
 \midrule[0.5pt]
FT & \ding[1.25]{55} & \ding[1.25]{55}  & \ding[1.25]{55} & \ding[1.25]{55} \\ \hline
NAD \cite{Li2021NeuralAD} & \ding[1.25]{51} & \ding[1.25]{55} & \ding[1.25]{55} & \ding[1.25]{55} \\ \hline
FT-SAM \cite{ft_sam_Zhu_2023_ICCV} & \ding[1.25]{51} & \ding[1.25]{51} & \ding[1.25]{55} & \ding[1.25]{55} \\ \hline
I-BAU \cite{Zeng2021AdversarialUO} & \ding[1.25]{51} & \ding[1.25]{51} & \ding[1.25]{51}  &\ding[1.25]{55}  \\ \hline
MCL \cite{Yue2023ModelContrastiveLF} & \ding[1.25]{51} & \ding[1.25]{51} & \ding[1.25]{55} & \ding[1.25]{51} \\
\midrule[0.5pt]
{\bf PGBD} (Ours)& \ding[1.25]{51} & \ding[1.25]{51} & \ding[1.25]{51} &  \ding[1.25]{51}\\ 
\bottomrule[1.2pt]
\end{tabular}%
}
\caption{Compared to naive fine-tuning (FT) and other post-hoc defenses, PGBD is {\em configurable} (able to utilize additional attack scenario information during defense), {\em scalable} (able to perform well across different types of attacks), {\em robust} (maintains performance with model and dataset changes), and {\em balanced} (balances the defense objectives of minimizing Attack Success Rate while retaining Clean Accuracy.}
\label{tab:model_sanitization_comp}
\end{table}
We briefly discuss backdoor data poisoning attacks, and their defense with a focus on model sanitization methods.
\Paragraph{Concept Based Model Improvement:}
\citet{Gupta2023ConceptDL} introduced \textit{concept distillation} for (de)sensitizing a model for a certain concept by moving model activations against (or towards) a particular CAV direction. CAV (Concept Activation Vector) indicates the activation space direction that points towards the location of a given concept \cite{tcav}. 
Recently, \citet{dong2024ocie} introduced Language-Guided CAV to utilize knowledge in CLIP and activation sample reweighing to enhance model correction by dynamically training with samples and aligning predictions with relevant concepts. CBMs \cite{koh2020concept} use manually defined vectors for supervision to train model to focus on certain concepts. Interactive methods like \cite{chauhan2023interactive, bontempelli2022concept, song2023img2tab} use user interaction to tune models for specific concepts. We utilize the debiasing capability of \citet{Gupta2023ConceptDL} in the backdoor setting with class-specific directions to define the trigger concept. 

\Paragraph{Attacks:} Based on the kind of trigger, backdoor data poisoning attacks are of three types:
\textit{(i)} \emph{Patch/Localized} trigger-based attacks use perturbations that alter only a small local region of the image \cite{8685687, Cheng_Liu_Ma_Zhang_2021, 10.1145/3319535.3354209}.
\textit{(ii)} \emph{Functional} trigger based variety of attacks perturb image globally, are generally imperceptible \cite{nguyen2021wanet, li2021anti} and do not require dataset label modification (\ie \textit{clean-label attack}) \cite{Barni2019ANB_signal, Liu2020Refool}. Dynamic backdoor attacks \cite{iab_nguyen2020input, dynamicba_salem2022dynamic, ssba_li2021invisible} can be either functional or patch-based and add a sample-level uniqueness constraint on the trigger by learning it as a function of both the image and the target class.
\textit{(iii)} \emph{Semantic} trigger based attacks use realistic triggers that naturally fit into the dataset scenes.
These attacks highlight real-world risk by poisoning the model with natural triggers in the deployment environment where the model is least protected. 
Face classification (with triggers like tatoos, sunglasses, hats \etc) is a common scenario used for such attacks \cite{Sarkar2022FaceHackAF, Chen2017TargetedBA, 9577800}.
We present a challenging semantic attack on faces and, for the first time, a successful defense against this attack type (\cref{tab:mainRes} ROF).
Additionally, we also defend against various other attack types with the same technique. Specifically, we defend against two patch based attacks \ie Badnet and Trojan; three functional trigger attacks \ie Blended (perceptible), Wanet (imperceptible) and Signal (clean-label) and three semantic attacks (sunglasses, tattoo and mask).

\Paragraph{Defenses:} Defense against backdoor attacks has been proposed in various settings and at different points in the training and inference pipelines.
Training Data Sanitization weeds out suspicious samples \textit{pre-training} \cite{Taheri@2020, 10.1007/978-3-030-13453-2_1, Tran2018SpectralSI, 10.1016/j.cose.2021.102280, pmlr-v139-hayase21a}. Robust training methods like Differential Privacy \cite{li2021anti, DBLP:conf/iclr/ChenL0Z21}, simple aggregation techniques like Bagging \cite{10.1007/978-3-642-21557-5_37, pmlr-v162-wang22m} and data augmentation \cite{borgnia2021strong, pmlr-v97-cohen19c} are also explored as potential defenses \textit{during training}. Some methods proposed for robust training \cite{certified_jia2021intrinsic}, and data sanitization \cite{certified_zeng2023towards} also provide algorithmically provable guarantees for their defense.
Full access to the complete training dataset is needed for the former while model parameters and training procedures are needed for the latter. Such defenses are expensive, impractical and even infeasible when training is outsourced to a third-party or when full access is not possible.

Test Data Sanitization \cite{chou2020sentinet, gao2019strip}, Model Inspection, (\ie detecting backdoored models) \cite{kolouri2020universal, xu2021detecting, chen2018detecting, certified_xiang2023cbd}, Trigger Reconstruction (\ie regenerating the perturbation by activation analysis) \cite{Wang2019NeuralCI, hu2021trigger, zhu2020gangsweep} and Model Sanitization (\ie finetuning/retraining using only a small clean trainset) \cite{Li2021NeuralAD,Yue2023ModelContrastiveLF, Zeng2021AdversarialUO} are some known \textit{post-training} defense tools. The former two do not alter the backdoored model and hence not useful for our goal.
Trigger synthesizers such as Neural Cleanse (NC, \cite{Wang2019NeuralCI}) and its variations TABOR \cite{TABOR} and AD\cite{xiang2020detection}, or alternatives like ABS \cite{10.1145/3319535.3363216}, though not designed for the purpose, can predict target class \textit{t}. The synthesized trigger might not fully match the original, but the target label prediction is reliable. NC also proposes a model sanitization by \emph{pruning neurons}, which shows a high response to trigger perturbed images. Recent neuron pruning works \cite{ANP_NEURIPS2021_8cbe9ce2, wu2022backdoorbench, pmlr-v202-li23v} build upon this and move away from the requirement of a synthesized trigger, but do not scale to multiple architectures.

While supervised learning-based tasks are the main focus of defense literature, there have been recent works that propose attacks and defenses for other paradigms of deep learning like reinforcement learning \cite{rl_chen2023bird, rl_wang2021backdoorl}, self-supervised learning \cite{ssl_feng2023detecting, ssl_li2023embarrassingly, ssl_saha2022backdoor, ssl_tejankar2023defending}, etc. While PGBD could scale to these paradigms, we restrict the scope of current work to supervised classification tasks.

\Paragraph{Discussion:} PGBD is a post-hoc model sanitization method. Previous works are done under different settings. The strongest assumes only the availability of a small ($\sim5\%$) subset of clean training data $D_S$ and the backdoored model $M_B$. Weaker setting additionally need the target class ($t$) to be known \cite{He2021DeepObliviateAP, backdoor_unlearning} and the weakest require the trigger used for the attack to be known \cite{Yue2023ModelContrastiveLF}. PGBD can be configured to work in all three settings. The base PGBD needs $D_S$ and $t$. Please note that the target can be inferred from $M_B$ using known techniques. Our ST-PGBD variant can take advantage of the known trigger prior. Our {\em no-target} variant NT-PGBD works in the strongest setting. See \cref{sec:variations} for a discussion on their relative merits and demerits. All PGBD variants work at least on-par with the best from the literature in their respective settings.

Early defenses scaled only to simple patch trigger based attacks \cite{Wang2019NeuralCI, Li2021NeuralAD} and couldn't scale to more sophisticated attacks. Recent defenses have scaled to functional and dynamic triggers \cite{Zeng2021AdversarialUO, ft_sam_Zhu_2023_ICCV, Yue2023ModelContrastiveLF}. However, no defense has been shown on semantic attacks and our adaptation of prior methods performed poorly. PGBD provides the first successful defense against semantic attacks (\cref{tab:mainRes} (ROF)). Finally, recent methods point out the lack of robustness of defenses, particularly when the percentage of poisoned training data during the attack was low \cite{min2023towards_fst}. Our experiments confirm this (\cref{tab:mainRes}) but PGBD was robust against the same. Overall, PGBD builds on the progress of defenses so far by scaling to previously unbeaten semantic attacks with improved robustness to changes in attack configurations (\cref{tab:model_sanitization_comp}). 

\section{Geometric Lens on Poisoned Activations}
\label{sec:validation}



Input samples are transformed into the successive activation spaces of network layers during classification. Geometric analysis of the activation spaces can give insights to understand and improve the model behavior in important ways. \citet{tcav} defined Concept Activation Vectors as an interpretability tool to understand the influence of different concepts using concept sets. \citet{Gupta2023ConceptDL} created debiased trained models on human interpretable concepts using an additional CAV loss. How can geometric manipulation defend against backdoor attacks? Geometrically, we observe that a poisoned sample will be (mis)directed towards the target class from the correct class. Penalizing the movement towards the target class in a fine-tuning step can post-hoc sanitize the poisoned model.

\begin{figure*}[t]
    \centering
    \includegraphics[width=0.7\textwidth]{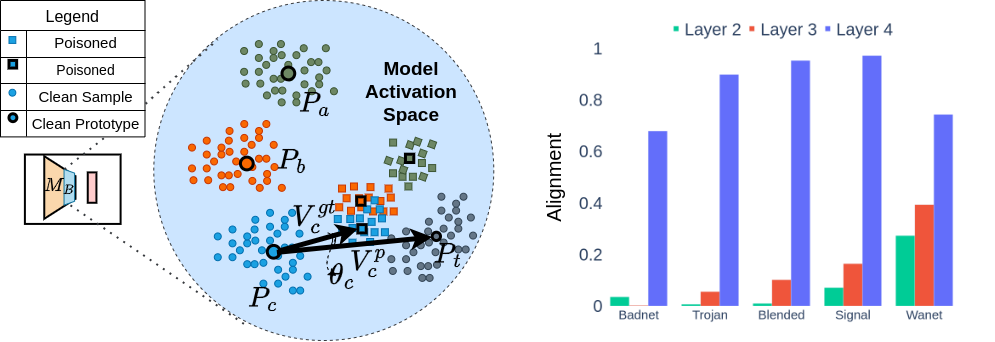}
    \caption{\textbf{[LEFT]} Visualization of $M_B$ activation space, with one of the clean class prototype ($P_c$) in blue and vectors ($V^p$ and $V^{gt}$) pointing to the target class prototype ($P_t$) and the poisoned prototype corresponding to class $c$. \textbf{[RIGHT]} Bar graph of alignment for the last three conv. layers of preActResNet18 model (Layer 4 denotes the last conv. layer). A value close to 1 indicates close alignment with the target class direction.}
  \label{fig:Validation}
\end{figure*}

\cref{fig:Validation} (left) shows clean and poisoned samples and their prototypes along with displacement vectors in a suitable activation space for different classes. In practice, the ground truth shift $V^{gt}_c$ is not known. The vector $V^P_c$ from class $c$ towards the target class $t$ can serve as a reasonable proxy, however. We call them Prototype Activation Vectors (PAV). \cref{fig:Validation} (right) shows that $V^{gt}_c$ and $V^P_c$ are well aligned in later layers of the network as their average cosine value over all classes $c \neq t$ is high. Some prior works \cite{Tran2018SpectralSI,Wang2019NeuralCI} assumed that poisoned samples cluster near the target class. We only assume a weaker, directional alignment. This is the geometric basis of our defense method that is explained next.

\begin{figure}
    \centering
    \includegraphics[width=0.9\linewidth, height=5cm]{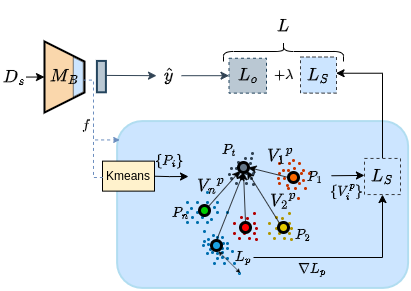}
  \caption{PGBD uses clean data $D_s$ to compute class prototypes. PAV $V^P_i$ for class $i$ points to target prototype $P_t$. Our new sanitization loss $L_S$ is the cosine distance of the PAV with the gradient of the corresponding prototype loss ($\nabla L_p$).}
  \label{fig:blockDiag}
\end{figure}

\section{Prototype Guided Backdoor Defense}
\label{sec:method}


Like prior methods, the basic PGBD system assumes the availability of the backdoored model $M_B$ and a small clean subset $D_s$ of the training data. The target class $t$ is also needed but is inferred using a method Neural Cleanse. (A more general variation tjat doesn't need $t$ is discussed in \cref{subsec:no_target}). The overall pipeline (\cref{fig:blockDiag}) for PGBD has two steps: (a) Calculating the class prototypes and Prototype Activation Vectors based on $t$ and (b) Finetuning $M_B$ on $D_s$ using an additional sanitization loss. We also use an optional module to map activations using a large pre-trained model for geometric reasoning in a richer space. We see this mapping retains the model accuracy better (\cref{fig:mapping_graph}).


\subsection{Estimating Prototype Activation Vectors}
\label{sec:method_prototype}

As discussed in \cref{sec:validation}, we leverage prototypes to perform class-specific geometric manipulation away from the target class. Prototypes are defined as the means of class activation clusters  \cite{keswani2022proto2proto}. With prototypes representing class clusters in the activation space, we propose Prototype Activation Vectors (PAVs) to define class specific directions of movement to be avoided during finetuning.

PAV (denoted by $V$) is the direction in activation space that points from one prototype towards another. 
For sanitizing $M_B$, we are interested in the direction from each class to the target class. We define {\em pure} PAV for class $c$ as
\begin{equation} 
\label{eq:pav}
    V^p_c = P_t - P_c.
\end{equation}
$P_t$ is the prototype for the target class, and $V^P_c$ and $P_c$ are the PAV and prototype for class $c$. Our base defense strategy PGBD uses this direction for loss calculation.

\subsection{Sanitizing the Model}
\label{sec:method_sanitization}

We finetune the poisoned model $M_B$ to sanitize it of its backdoor. Finetuning follows the original training procedure but with an additional {\em sanitization loss} $L_S$ introduced at a chosen layer.

\Paragraph{Losses:} 
Sanitization has two objectives: preserve the original Clean Accuracy (CA) for clean samples and reduce Attack Success Rate (ASR) for poisoned samples. Thus, a poisoned sample $d$ originally belonging to class $c$ will be assigned the correct label $c$ post sanitization instead of the incorrect target class label $t$.

We introduce a novel loss term while finetuning. For a data item $(x,c) \in D_s$, we calculate prototype loss $L_p$ denoting the Euclidean distance of $f(x)$ in a given layer from its correct class prototype $P_c$. Note that compared to the clean ones, $L_p$ is expected to be high for poisoned samples.
To discourage movement towards the target prototype, we penalize the contribution of loss gradient in the direction of PAV.
Similar to \citet{Gupta2023ConceptDL}, we use cosine similarity between the prototype loss gradient $\nabla L_p$ and PAV $V$ (if base PGBD is being used, $V = V^p$) as the {\em sanitization loss} $L_s$. 
Intuitively, $\nabla L_p$ indicate shift required to bring a sample closer to its class prototype while $L_s$ ensures movement restriction towards direction $V$.
The final loss for the sanitization step is $L = L_o + \lambda L_s$, where $L_o$ is the original classification loss.
\begin{align}
\small
L_p    &=   \norm{f(x) - P_{c} }^2  &// \rm{MSE} 
\label{eqn:Lp} \\
L_s &= (\nabla L_p \boldsymbol{\cdot} V) / (\norm{\nabla L_p} \norm{V}) &\rm{// projection}
\label{eqn:Ls} \\
L &= L_o + \lambda L_s. & \label{eqn:L}
\end{align}
Note that $L_o$ (usually a cross-entropy loss) is computed on the final linear layer outputs, whereas $L_p$ and $L_s$ are computed in an intermediate activation space. We use the last convolutional layer based on the observations from \Cref{sec:validation}, but other layers could also be used. Overall, our finetuning loss penalizes directional movement to target class using $L_s$ and penalizes clustering away from the clean class prototype using $L_p$. The impact of $\lambda$ is discussed in \cref{subsec:ablations}.

\Paragraph{Large-Model Mapping:}
The training data used by most practical systems come from a relatively narrow distribution compared to the space of all images. The geometric insights derived from there may hence be limited. 
Feature spaces of large pre-trained models are known to possess richer geometric and algebraic properties as they see huge data covering a broader region of the space of images \cite{caron2021emerging}. These large models can be harnessed effectively to enrich the feature space by mapping it to their space as done by \citet{Gupta2023ConceptDL}. We adapt their method to map the prototype vectors to the space of a large model and compute PAVs in its space.
Specifically, we \emph{lift} the prototypes from the ($M_B$) activation space to the space of the pre-trained model with the help of a lightweight mapping module.
The mapping module is a simple reversible linear transform implemented using a shallow autoencoder. It can be quickly trained in a self-supervised manner by mapping the features from the teacher (i.e., large model) space to the student space and back without requiring any ground truth. Importantly, once we store the large models features on $D_s$, we no longer require the large model as the mapping module only requires the features. The large models are just a tool to get enhanced performance. Several large pre-trained vision models are readily available. We primarily use DINOv1 \cite{caron2021emerging} in this work.
\section{Experiments \& Results} 
\begin{table*}[t]
\footnotesize
\centering
\begin{tabular}{c|l||cc||ccc|ccc|ccc|ccc|ccc||ccc}
\toprule[1.2pt]
  \multicolumn{2}{c||}{\bf Method \faCaretRight} &
  \multicolumn{2}{c||}{\bf Baseline} &
  \multicolumn{3}{c|}{\bf FT} &
  \multicolumn{3}{c|}{\bf NAD} &
  \multicolumn{3}{c|}{\bf I-BAU} &
  \multicolumn{3}{c|}{\bf FT-SAM} &
  \multicolumn{3}{c||}{\bf MCL} &
  \multicolumn{3}{c}{\bf PGBD} \\
\hline 
 & Attack & CA  & ASR  & CA  & ASR  & $\Gamma$ & CA  & ASR  &  $\Gamma$ & CA  & ASR  & $\Gamma$ & CA  &  ASR & $\Gamma$ & CA  & ASR  & $\Gamma$ & CA  & ASR  & $\Gamma\uparrow$  \\
\midrule[0.5pt]
\multirow{5}{*}{\STAB{\rotatebox[origin=c]{90}{CIFAR10}}}
& Badnet  & 92.34 & 88.93  & 92.66 & 8.43  & 0.95 & 92.64 & 4.92  & \underline{0.97} & 89.56 & 1.81  & \underline{0.97} & 92.26 & 1.14  & \textbf{0.99} & 89.76 & 0.01 & \textbf{0.99} & 90.66 & 0.82 & \textbf{0.99} \\
& Trojan  & 93.00 & 100.00 & 93.30 & 99.92 & 0.50 & 93.07 & 99.76 & 0.50       & 63.85 & 10.42 & 0.79       & 93.04 & 88.12 & 0.56          & 78.43 & 8.51 & \underline{0.88}    & 83.60 & 6.76 & \textbf{0.92} \\
& Blended & 93.06 & 92.94  & 93.47 & 92.40 & 0.50 & 80.79 & 3.84  & 0.91       & 81.54 & 2.31  & 0.93       & 93.03 & 49.67 & 0.73          & 90.28 & 2.18 & \textbf{0.97} & 86.11 & 4.87 & \underline{0.94}    \\
& Sig     & 92.90 & 89.04  & 93.51 & 83.87 & 0.53 & 81.79 & 6.81  & 0.90       & 89.23 & 18.03 & 0.88       & 93.17 & 37.50 & 0.79          & 80.71 & 4.51 & \underline{0.91}    & 87.18 & 0.31 & \textbf{0.97} \\
& Wanet   & 89.98 & 97.60  & 93.39 & 18.37 & 0.91 & 93.32 & 10.87 & 0.94       & 90.90 & 1.30  & \underline{0.99} & 93.68 & 0.12  & \textbf{1.00} & 86.08 & 3.64 & 0.96          & 88.54 & 2.36 & 0.98 \\ 
& IAB & 90.49 & 91.01 & 93.05 & 84.44 & 0.54 & 92.99 & 80.51 & 0.56 & 91.48 & 10.23 & 0.94 & 93.05 & 8.02 & \underline{0.96} & 80.38 & 0.11 & 0.94 & 89.43 & 2.68 & \textbf{0.98} \\
\hline 
\multirow{3}{*}{\STAB{\rotatebox[origin=c]{90}{ROF}}}
& Sunglass & 93.33 & 86.19 & 98.33 & 74.45 & 0.57 & 70.39 & 40.41 & 0.64 & 90.53 & 82.87 & 0.50 & 89.35 & 25.97 & \underline{0.83} & 33.33 & 38.12 & 0.46 & 71.67 & 4.97 & \textbf{0.86} \\
& Tattoo     & 78.40 & 72.10 & 92.31 & 31.85 & 0.78 & 84.60 & 18.46 & 0.87 & 82.91 & 23.64 & 0.84 & 92.30 & 9.23  & \underline{0.94} & 40.38 & 2.38  & 0.74 & 86.53 & 2.40 & \textbf{0.98} \\
& Mask       & 69.23 & 99.69 & 73.07 & 21.10 & 0.89 & 28.12 & 0.00  & 0.70 & 72.69 & 59.03 & 0.70 & 75.90 & 48.19 & \underline{0.76} & 21.10 & 9.60  & 0.60 & 63.46 & 3.33 & \textbf{0.94} \\
\hline 
\multirow{4}{*}{\STAB{\rotatebox[origin=c]{90}{CIFAR100}}} 
& Badnet  & 67.32 & 86.98  & 66.82 & 0.43  & 0.99 & 66.72 & 0.01  & \textbf{1.00} & 61.20 & 0.07 & 0.95          & 64.98 & 0.81  & \underline{0.98}    & 47.65 & 0.00 & 0.85 & 64.29 & 0.01 & \underline{0.98}    \\
& Trojan  & 70.02 & 100.00 & 68.93 & 99.40 & 0.50 & 68.34 & 89.42 & 0.54          & 66.00 & 0.89 & \textbf{0.97} & 65.50 & 84.11 & 0.55          & 31.50 & 0.00 & 0.72 & 62.50 & 0.02 & \underline{0.95}    \\
& Blended & 69.01 & 99.48  & 67.63 & 97.04 & 0.50 & 67.79 & 97.80 & 0.50          & 61.54 & 0.35 & \underline{0.94}    & 64.92 & 84.48 & 0.55          & 28.03 & 0.00 & 0.70 & 61.44 & 0.00 & \textbf{0.95} \\
& Wanet   & 63.84 & 91.47  & 68.53 & 0.57  & 1.00 & 68.85 & 1.93  & 0.99          & 63.59 & 7.12 & 0.96          & 67.67 & 1.71  & \textbf{0.99} & 42.58 & 0.00 & 0.83 & 62.34 & 0.66 & \underline{0.98} \\
\hline 
\multirow{4}{*}{\STAB{\rotatebox[origin=c]{90}{GTSRB}}} 
& Badnet  & 96.61 & 83.86  & 97.97 & 51.80  & 0.69       & 97.76 & 57.30 & 0.66          & 96.85 & 0.00  & 1.00          & 98.30 & 1.70  & 0.99          & 92.03 & 0.00  & 0.98 & 97.26 & 0.00 & \textbf{1.00} \\
& Trojan  & 98.17 & 100.00 & 98.70 & 100.00 & 0.50       & 96.93 & 0.17  & \textbf{0.99} & 93.45 & 2.70  & 0.96          & 98.10 & 4.31  & \underline{0.98}    & 82.12 & 0.00  & 0.92 & 96.50 & 0.11 & \textbf{0.99} \\
& Blended & 98.66 & 96.33  & 98.51 & 94.86  & 0.51       & 96.44 & 35.61 & 0.80          & 86.25 & 17.29 & 0.85          & 98.10 & 19.70 & \underline{0.89}    & 33.75 & 0.02  & 0.67 & 86.20 & 0.72 & \textbf{0.93} \\
& Wanet   & 98.04 & 82.14  & 99.24 & 31.70  & \underline{0.81} & 99.15 & 41.55 & 0.75          & 95.95 & 0.05  & \textbf{0.99} & 99.30 & 1.77  & \textbf{0.99} & 79.90 & 29.31 & 0.73 & 97.14 & 0.33 & \textbf{0.99}\\
\hline 
\multirow{3}{*}{\STAB{\rotatebox[origin=c]{90}{TINY}}} 
& Badnet  & 57.07 & 94.92 & 58.14 & 89.14 & 0.53 & 40.94 & 65.55 & 0.51 & 49.59 & 74.08 & 0.54 &  52.85  & 67.50 & 0.61 & 30.41 & 0.00  & \underline{0.77} & 48.46 & 18.39 & \textbf{0.83} \\
& Trojan  & 56.94 & 98.56 & 55.72 & 97.40 & 0.50 & 38.52 & 90.11 & 0.38 & 48.42 & 86.10 & 0.49 &  52.8  & 98.89  & 0.46 & 20.39 & 0.00  & \underline{0.68} & 41.22 & 21.05 & \textbf{0.76} \\
& Blended & 57.04 & 95.59 & 55.83 & 89.11 & 0.52 & 29.66 & 78.14 & 0.35 & 49.25 & 63.77 & 0.60 & 52.39 & 92.68 & 0.47 & 32.30 & 16.58 & \underline{0.70} & 42.00 & 14.09 & \textbf{0.79} \\
\hline\hline 
\multirow{8}{*}{\STAB{\rotatebox[origin=c]{90}{MEAN}}} 
& Badnet & 78.33 & 88.67 & 78.90 & 37.45 & 0.79 & 74.52 & 31.95 & 0.78 & 74.30 & 18.99 & 0.87 & 77.10 & 17.79 & 0.89 & 64.96 & 0.00 & \underline{0.90} & 75.17 & 4.81 & \textbf{0.95} \\
& Trojan & 79.53 & 99.64 & 79.16 & 99.18 & 0.50 & 74.22 & 69.86 & 0.60 & 67.93 & 25.03 & \underline{0.80} & 77.36 & 68.86 & 0.64 & 53.11 & 2.13 & \underline{0.80} & 70.96 & 6.99 & \textbf{0.90} \\
& Blended & 79.44 & 96.08 & 78.86 & 93.35 & 0.51 & 68.67 & 53.85 & 0.64 & 69.65 & 20.93 & \underline{0.83} & 77.11 & 61.63 & 0.66 & 46.09 & 4.70 & 0.76 & 68.94 & 4.92 & \textbf{0.90} \\
& Sig        & 92.90 & 89.04  & 93.51 & 83.87 & 0.53 & 81.79 & 6.81  & 0.90 & 89.23 & 18.03 & 0.88       & 93.17 & 37.50 & 0.79          & 80.71 & 4.51  & \underline{0.91} & 87.18 & 0.31 & \textbf{0.97} \\
& Wanet      & 83.95 & 90.40  & 87.05 & 16.88 & 0.90 & 87.11 & 18.12 & 0.89 & 83.48 & 2.82  & \underline{0.98} & 86.88 & 1.20  & \textbf{0.99} & 69.52 & 10.98 & 0.84       & 81.77 & 0.90 & \underline{0.98}    \\
& Sunglass & 93.33 & 86.19  & 98.33 & 74.45 & 0.57 & 70.39 & 40.41 & 0.64 & 90.53 & 82.87 & 0.50       & 89.35 & 25.97 & \underline{0.83}    & 33.33 & 38.12 & 0.46       & 71.67 & 4.97 & \textbf{0.86} \\
& Tattoo     & 78.40 & 72.10  & 92.31 & 31.85 & 0.78 & 84.60 & 18.46 & 0.87 & 82.91 & 23.64 & 0.84       & 92.30 & 9.23  & \underline{0.94}    & 40.38 & 2.38  & 0.74       & 86.53 & 2.40 & \textbf{0.98} \\
& Mask       & 69.23 & 99.69  & 73.07 & 21.10 & 0.89 & 28.12 & 0.00  & 0.70 & 72.69 & 59.03 & 0.70       & 75.90 & 48.19 & \underline{0.76}    & 21.10 & 9.60  & 0.60       & 63.46 & 3.33 & \textbf{0.94} \\
\bottomrule[1.2pt]
\end{tabular}%
\caption{Quantitative comparison between five different defenses (Finetuning, NAD \cite{Li2021NeuralAD}, I-BAU\cite{Zeng2021AdversarialUO}, FT-SAM\cite{ft_sam_Zhu_2023_ICCV}, MCL\cite{Yue2023ModelContrastiveLF}) and our base PGBD method for five benchmarks (CIFAR10, ROF, CIFAR100, GTSRB, TinyImagenet). We report three metrics (CA $\uparrow$, ASR $\downarrow$, and $\Gamma \uparrow$) for each of the five attack types (Badnet \cite{8685687}, Trojan \cite{Cheng_Liu_Ma_Zhang_2021}, Blended \cite{Chen2017TargetedBA}, Signal \cite{Barni2019ANB_signal}, Wanet \cite{nguyen2021wanet}, and IAB \cite{iab_nguyen2020input}) and three semantic attack situations (Sunglass, Mask, Tatttoo). The best and second best values are in \textbf{bold} and \underline{underline}, respectively. Overall, PGBD achieves the best DEM($\Gamma$) across all attacks when averaged across datasets.}
\label{tab:mainRes}
\end{table*}
\label{sec:results}
\label{sec:res_expConfig}

\Paragraph{Setup:} We implement our approach using PyTorch \cite{pytorch} on a single 12 GB Nvidia 2080Ti GPU.
We use a poisoning rate of 1\% for CIFAR10, GTSRB, and TinyImagenet datasets and 10\% for CIFAR100 to ensure satisfactory ASR. 
All our experiments use SGD optimizer with constant values for learning rate=0.0001, momentum=0.9 and weight decay=0.0001.
During defense, we finetune for 35 epochs without dropout or learning rate decay with an average time of 35-45 seconds per epoch.
Fixed hyperparameters ($\lambda=10$ and $\alpha=0.75$) via manual grid search are used for all experiments unless stated otherwise. 

\Paragraph{Models:} We employ preact-ResNet18 \cite{he2016identity} as model architecture for our student and other defenses for all our experiments.
We use the average of three centroids obtained using Kmeans for each class prototype. 
When using large-model mapping, we use the DINOv1 model \cite{caron2021emerging} with ViT-Base8 architecture and 384 dimensional features extractor encoder implementation \cite{tschernezki22neural} as the teacher space. We train the mapping module for 5 epochs.
Please see the Appendix for more details (\cref{sec:supple_implementation}).

\Paragraph{Defense Efficacy Measure:}
Clean accuracy (CA) and attack success rate (ASR) are the standard performance metrics for defense. A successful backdoor attack results in poisoned models with high accuracy on clean samples (CA) and high success (ASR) by misclassifying the test samples with triggers as the target class.
A perfect defense will retain the CA values of the baseline while driving the ASR to 0. A trade-off between CA and ASR can be observed in practice.
We propose {\em Defense Efficacy Measure} (DEM) considering this trade-off. 

Let CA$_P$, ASR$_P$ and CA$_D$, ASR$_D$ be the respective CA and ASR values of the poisoned baseline and post defense. Consider:
\begin{align}
\small
    \Delta C &= \frac{\mathrm{CA}_P - \mathrm{CA}_D}{\mathrm{CA}_P}; \; \quad
    \Delta A = \frac{\mathrm{ASR}_P - \mathrm{ASR}_D}{\mathrm{ASR}_P} \nonumber \\
    \delta_C &= 1 - \max(\Delta C,0); \; \quad  \delta_A = \max(\Delta A,0) \\
    \Gamma &= \frac{1}{2}(\delta_C + \delta_A).
\label{eqn:DEM_linear}     
\end{align}
Here, $\Delta C$ and $\Delta A$ are the additive inverse of change in CA and ASR \wrt baseline. $\delta_C$ and $\delta_A$ are linearized values and will be 1 for a perfect defense.
The DEM $\Gamma$ is their mean which is 1 for a perfect defense and 0 for a poor one.
Note that \citet{ft_sam_Zhu_2023_ICCV} also propose a similar metric but do not use the individual $\delta_C$ and $\delta_A$ terms.

\begin{figure}
    \centering
    \includegraphics[width=\linewidth]{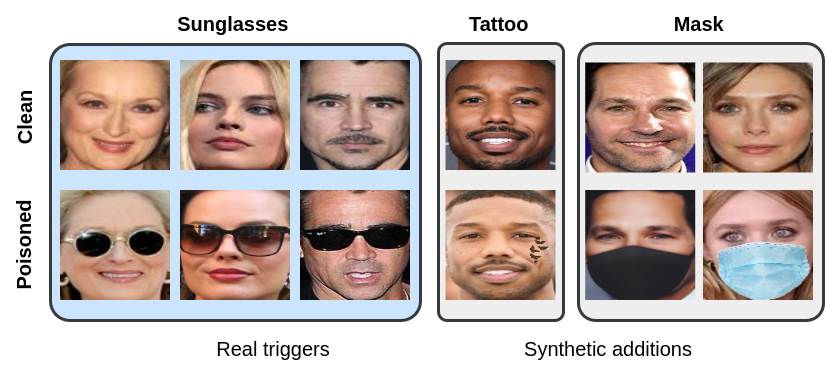}
    \caption{Our proposed face occlusion semantic attack benchmark using sunglasses, tattoos, and masks as triggers.}
    \label{fig:semantic_teaser}
\end{figure}

\Paragraph{Semantic Backdoor Attack:}
Semantic backdoors use an inconspicuous scene object as a trigger and are easy to carry out during inference. As far as we know, no defense method has been proposed for this attack category till date. Successful semantic attacks with real world face occlusions have been reported earlier \cite{Chen2017TargetedBA, 10.1145/3393527.3393567} but their data is not public. 
Hence we create our own realistic semantic attack dataset for face recognition similar to \cite{Sarkar2022FaceHackAF, Chen2017TargetedBA, 10.1145/3393527.3393567}.
We create a face occlusion attack using the real-world occluded faces public dataset (ROF) \cite{erakiotan2021recognizing}.
ROF consists of 5559 images of 180 celebrities. All celebrities have images with sunglasses which we directly use as the poisoned dataset.
For synthetic variants, we use Snapchat\footnote{https://web.snapchat.com/} filters to create poisoned datasets. We use a single tattoo filter for the tattoo-based attack and multiple mask filters for the mask-based attack (see \Cref{fig:semantic_teaser}). We filter out 10 classes for each trigger ranked by the number of neutral images (for masks and tattoo triggers) and the number of occluded images (for sunglasses triggers). We use ResNet50 architecture for training on this attack, given the larger image size.

\paragraph{Comparison:}
As observed from the last few rows (MEAN) in \Cref{tab:mainRes}, we achieve state-of-the-art performance in DEM for all attacks averaged across four datasets (CIFAR10 \cite{krizhevsky2009learning}(10 lasses), ROF \cite{erakiotan2021recognizing}(10 classes), CIFAR100 \cite{krizhevsky2009learning}(100 classes), GTSRB \cite{STALLKAMP2012323}(43 classes), TinyImagenet \cite{le2015tiny}(200 classes)).
For ASR, apart from Badnet and Blended where we are second best, our method shows maximum ASR reduction among all the defenses.
Specifically, we achieve 95\%, 93.2\%, 95\%, 99.9\%, 99.6\%, and 97.3\% average $\Delta$A for all the attacks respectively. We achieve an overall average $\Delta$C of just 7\% over all datasets and attacks, displaying the balance in our defense strategy.
We achieve state-of-the-art performance on Signal, an attack which previous works have struggled to defend against\cite{Yue2023ModelContrastiveLF, ft_sam_Zhu_2023_ICCV}. 
In CIFAR100, NAD achieves better CA at the cost of ASR and  low DEM values, highlighting the utility of our metric. With respect to overall consistency in DEM values, I-BAU comes second best, followed by MCL, FT-SAM, NAD, and FT, respectively.
The effect of low poisoning rate is especially visible with the TinyImagenet dataset (\cref{tab:mainRes}(TINY)), where the results are considerably off from the corresponding self-reported 10\% poisoning rate results for the defenses. While dataset-level performance is the least here, PGBD maintains robustness and state-of-the-art performance even in this case. 
Also, basic finetuning-based defense (FT) performs well for the weak Badnet and Wanet attacks, supporting our previous observations in \Cref{sec:validation}.
PGBD outperforms MCL in CA retention while performing at least on-par in ASR, which shows that our sanitization technique is more precise than the sample contrastive loss. Overall, we achieve the best or close second best DEM values across all attacks over all datasets and achieve on par or better ASR reduction($\Delta A$) while safeguarding CA, clearly showing the scalability and robustness of our method.


Our PGBD method easily extends to semantic attacks unlike previous literature, as seen in the results for ROF dataset in \Cref{tab:mainRes}. We see PGBD is the only defense that achieves consistent performance across all three triggers with an average DEM of 0.93 while the next best is FT-SAM with an average DEM of 0.84, and the rest of the defenses at much lower values. We achieve greater ASR reduction than FT-SAM, while FT-SAM achieves most of the DEM through CA retention. It is important to note that all previous works struggle to reduce ASR in attacks where there is variance in triggers (sunglasses and/or mask) but perform relatively better in single trigger case (tattoo), highlighting a key vulnerability in existing defense designs that we tackle.
Overall, PGBD scales to all triggers while balancing CA retention and ASR reduction.

\Paragraph{Defense against adaptive attacks:}
Adaptive attacks attempt to camouflage poisoned samples into the corresponding clean samples while retaining attack potency (ASR). Defenses that rely on poisoned samples clustering together and away from the corresponding clean samples fail against these attacks. We show results against a recent adaptive attack \cite{xia2022enhancing_mmdr} using the Badnet trigger in \Cref{tab:reb_mlmmdr}. Results confirm the importance and robustness of our directional objective in \Cref{eqn:Ls}. The superior CA retention also reflects the benefits of PGBD, which relies only on the target class direction in the activation space, as opposed to MCL, which relies on synthesized triggers.
\begin{table}
\small
\begin{tabular}{l||cc||ccc|ccc}
\toprule[1.2pt]
 & \multicolumn{2}{c||}{\textbf{Baseline}} & \multicolumn{3}{c|}{\textbf{MCL}}      & \multicolumn{3}{c}{\textbf{PGBD}} \\
Attack & CA  & ASR  & CA  & ASR  & $\Gamma$ & CA  & ASR & $\Gamma$  \\
\midrule[1.5pt]
Badnet*  & 88.17 & 98.08  & 74.06  & 1.53  & 0.91 & 88.6  & 0.34 & \textbf{0.99} \\
\bottomrule[1.2pt]
\end{tabular}
\caption{Results on ML MMDR adaptive attack with Badnet (denoted as Badnet*) on CIFAR10 dataset.}
\label{tab:reb_mlmmdr}
\end{table}
\begin{figure}[b]
    \centering
    \includegraphics[width=\linewidth]{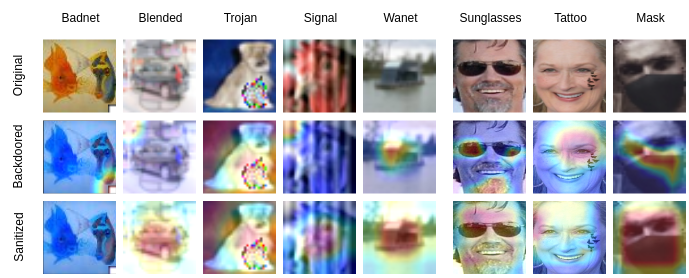}
    \caption{GradCAM visualizations before and after applying PGBD. Initially, the model focuses on backdoor triggers (red regions). Post-PGBD, the focus shifts to relevant class features, regaining model utility and robustness. }
    \label{fig: gradcam}
\end{figure}
\paragraph{GradCAM visualizations:}
We compare GradCAM \cite{gradcam} visualization of the last convolutional layer of $M_B$ on poisoned images from all the attacks in \Cref{fig: gradcam} before and after PGBD. In patch-based attacks, we observe that the $M_B$ focuses on the region where the trigger is (bottom right in the Badnet and Trojan columns). On the other hand, functional trigger-based attacks seem to cause learning of irrelevant features (top-left in the Signal and Blended columns). In the case of semantic attacks, the model appears to have learned the corresponding trigger feature of sunglasses, tattoo, or masks. Post-defense using PGBD, we consistently observe a complete focus on the relevant features of the subject of the image, displaying successful erasure of misclassifying features.
\section{Variations}
\label{sec:variations}
We propose two new variations of PGBD in this section: \textit{(i)} ST-PBGD that uses a synthesized trigger from \cite{Tao2022BetterTI} to generate synthetic PAVs ($V^s$), and \textit{(ii)} \textit{no target} or NT-PGBD, where we modify the PGBD pipeline to work without knowledge of target label (t). We show results on CIFAR10 for all three variations in \Cref{tab:variations_mean}.
\Paragraph{ST-PGBD:}
\label{subsec:PGBD_synth}
In this formulation, we estimate PAVs by using the trigger obtained from trigger synthesis methods. Note these are same requirements as MCL. We define a \textit{synthetic} PAV per class represented by $V^s_c$ where, 
$V^s_c = P'_c - P_c$.
Here \(P'_c\) is the class-wise prototype calculated using activations of all \(x' \in D'_s\), where \(D'_s\) is obtained by adding the synthesized trigger to \(D_s\). 
We observe better ASR reduction with $V^s$ for Trojan and Wanet attacks (\Cref{tab:variations_mean}), but $V^p$ performs better overall in terms of CA retention and DEM scores. The poor performance on semantic attacks can be attributed to the dependence on trigger synthesis (we use \cite{Tao2022BetterTI}). Importantly, ST-PGBD outperforms MCL across all attacks. We provide additional insights along with comparisons over all datasets in the Appendix (\cref{tab:abl_variations}).
\Paragraph{NT-PGBD:}
\label{subsec:no_target}
We observe from \Cref{eqn:L} that instead of using a singular target label $t$, iterating over all labels simply ensures prototypes are well separated and does not lead to performance degradation.
This suggests that cycling through all classes while treating the current class as the target should result in a still more robust backdoor defense.
We show results for the \textit{no-target} variation NT-PGBD in \Cref{tab:variations_mean} with cycling intervals of three, four, one, and one epoch for CIFAR10, ROF, GTSRB, and CIFAR100, respectively. 

We observe close performance to $V^p$ and better performance than $V^s$ overall. Specifically, when compared with I-BAU (the state of the art in the target label independent methods), NT-PGBD achieves an average DEM of 0.93 \vs 0.83 of I-BAU's. However, due to a minimum cycling time requirement of 1 epoch, NT-PGBD would need to be run for longer on larger datasets (for example, 30 epochs on CIFAR10 but 100 epochs on CIFAR100). 

As a solution, this variant can also be combined with trigger synthesis literature (like NC \cite{Wang2019NeuralCI}), where we use top $k$ target label predictions instead of all to reduce the computational load.
Overall, we recommend NT-PGBD as a foolproof defense if there is low confidence in the predicted target label or when the target selection of the attack is suspected to be arbitrary. In all other cases, we suggest using the base PGBD.


\begin{table*}

\footnotesize
\centering
\begin{tabular}{@{}L{12mm}||C{8mm}C{8mm}||C{8mm}C{8mm}C{8mm}|C{8mm}C{8mm}C{8mm}|C{8mm}C{8mm}C{8mm}@{}}
\toprule[1.2pt]

Method \faCaretRight & \multicolumn{2}{c||}{Baseline} & \multicolumn{3}{c|}{NT-PGBD} & \multicolumn{3}{c|}{ST-PGBD} & \multicolumn{3}{c}{PGBD} \\
Attack \faCaretDown & CA & ASR & CA & ASR & $\Gamma$ & CA & ASR & $\Gamma$ & CA & ASR & $\Gamma\uparrow$\\
\hline 
Badnet     & 85.42 & 86.59                     & 81.11 & 0.32  & 0.97          & 83.60 & 0.58  & \underline{0.98} & 83.74 & 0.28 & \textbf{0.99} \\
Trojan     & 87.06 & 100.0                     & 77.32 & 1.47  & \underline{0.93}    & 76.02 & 0.93  & \underline{0.93} & 80.04 & 2.29 & \textbf{0.95} \\
Blended    & 86.89 & 96.24                     & 78.41 & 0.39  & \textbf{0.94} & 73.45 & 2.76  & 0.90       & 76.93 & 1.88 & \underline{0.93}    \\
Sig        & 92.90 & 89.04                     & 90.06 & 0.02  & \textbf{0.98} & 89.94 & 7.51  & 0.94       & 87.18 & 0.31 & \underline{0.97}    \\
Wanet      & 83.95 & 90.40                     & 81.51 & 0.91  & \textbf{0.98} & 79.58 & 0.51  & \underline{0.97} & 81.77 & 0.90 & \textbf{0.98} \\
Sunglass & 93.33 & 86.19 & 63.33 & 1.67  & \underline{0.83}    & 59.07 & 0.85  & 0.81       & 71.67 & 4.97 & \textbf{0.86} \\
Tattoo     & 78.84 & 72.1                      & 74.33 & 3.82  & \underline{0.94}    & 48.86 & 1.3   & 0.80       & 86.53 & 2.40 & \textbf{0.98} \\
Mask       & 69.23 & 96.66                     & 53.97 & 0.414 & \underline{0.89}    & 76.92 & 56.67 & 0.71       & 63.46 & 3.33 & \textbf{0.94} \\
\bottomrule[1pt]
\end{tabular}

\caption{Mean results over all datasets for the three variants of PGBD. All variants perform better than SOTA in their respective settings. The performance order among the three is: PGBD $>$ NT-PGBD $>$ ST-PGBD}
\label{tab:variations_mean}
\end{table*}
\section{Ablations}
\label{subsec:ablations}
We study different defense and attack scenarios to test the robustness of our method. We also ablate on design choices here for PGBD and ST-PGBD. The quantitative results appear in the Appendix File for all ablations except large-model mapping. We summarize the important observations in the main paper.
\begin{figure}
    \centering
    \includegraphics[width=\linewidth, height=37mm]{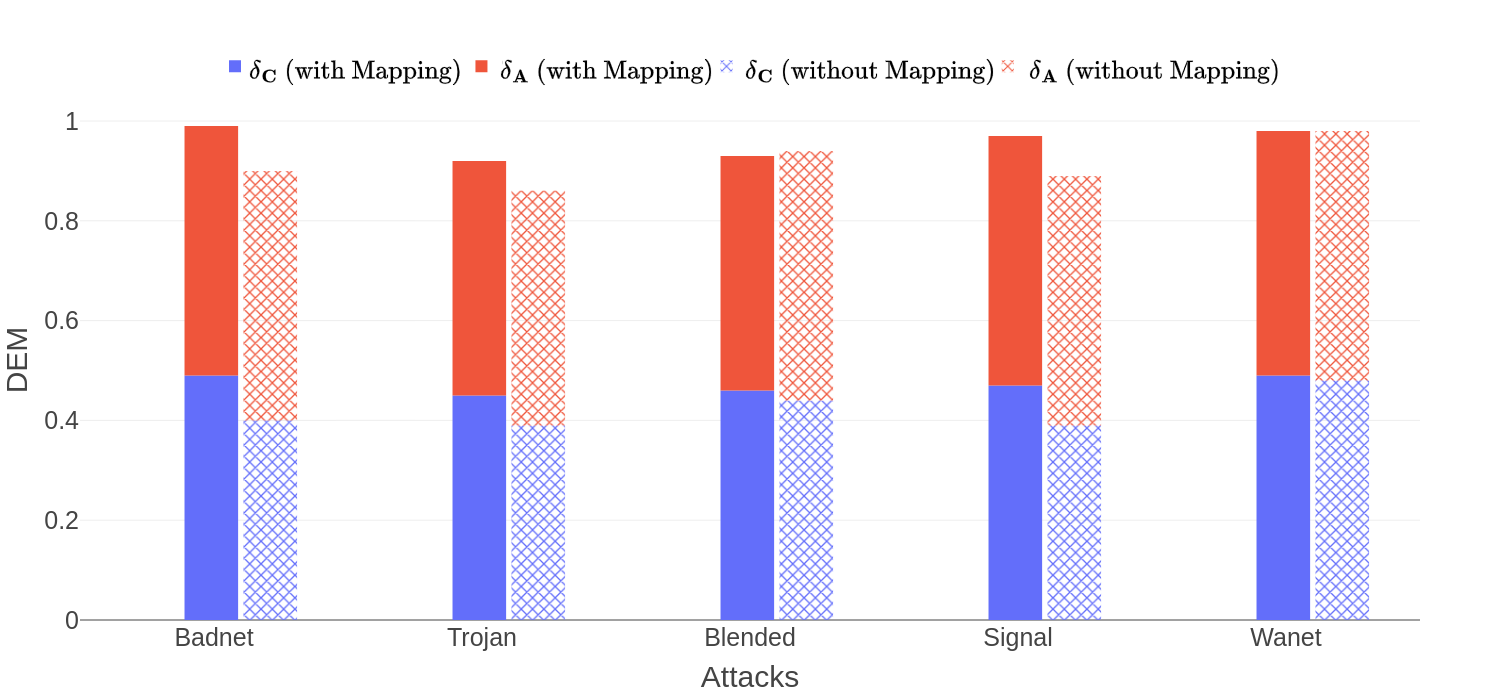}
    \caption{Comparision of PGBD with and without large-model mapping across 5 attacks on CIFAR10. Mapping (block bars) aids in CA retention (higher $\delta_C$) at the cost of slightly lower ASR reduction as compared to the no mapping case (patterned bars). Refer to \Cref{eqn:DEM_linear} for $\delta_C$ and $\delta_A$ definitions.}
    \label{fig:mapping_graph}
\end{figure}

\Paragraph{Effect of Large-Model Mapping:}
\Cref{fig:mapping_graph} shows results with and without mapping on the main PGBD variants. 
We observe an overall decrease in CA without mapping for both types of PAVs. This supports the intuition that large foundational models enriches the activation space \cite{Gupta2023ConceptDL}. Overall, DEM scores almost always improve with mapping. We recommend skipping large-model mapping only when ASR reduction is a very high priority. We look at the effect of mapping on the $V^s$ and {no target} variants in the Appendix material.

\Paragraph{Attack-time Ablations}
We test the robustness of PGBD against changes in attack time parameters: target label, poisoning rate (the subset of training data that is poisoned during attack time), target class in the form of {\bf all-to-all} attacks (where target class of $i$ is set to $(i+1)$ modulo the number of classes, \cref{tab:abl_all2all}), model size, model architecture, and the size of the datasets. PGBD maintains good defense performance with any target label and poisoning rate even up to 20\%. 
NT-PGBD achieves a mean $\Gamma$ of 0.96 on all-to-all, which shows potential to use it as an all-out defense irrespective of target class. PGBD demonstrates robustness with alternative model sizes and architectures, such as VGG19. Please see the Appendix file for details.

\Paragraph{Defense-time Ablations}
We vary the size of available clean data ($D_s$) and observe no adverse effect on CA retention and nearly uniform ASR reduction. This demonstrates PGBD's robustness. In comparison, MCL faces CA drop of $>=$20\% in a similar situation. We also study the impact of $\lambda$ (\Cref{eqn:L} and find that an optimum value of 10 balances the sanitization goal and the task objective. Small $\lambda$ maintains CA but fails to reduce ASR and vice-versa. We also vary the hyperparameter updation rate ($\alpha$) that controls PAV estimation during finetuning. We also try out an alpha scheduler based on the observation that most of the defense is achieved in the initial epochs. We observed slightly improved performance on Badnet, Blended, and Trojan attacks but consistent behaviour overall.

\Paragraph{Design Ablations:}
We explored other design choices of PGBD \cref{sec:ablations_supple}, such as the prototype computation process, the large-model used for mapping, the mapping process, and the use of ground truth triggers for PGBD $V^s$ instead of synthesized triggers. Overall, our analysis establishes the versatility of PGBD as a post-hoc defense while empirically validating prior insights.




\section{Conclusions}
We present a new defense strategy against backdoor attacks leveraging geometric configuration of class prototypes. Our PGBD method exhibits overall best performance for multiple attacks over several datasets, with and without the prior knowledge of the target class. We showed the first-ever defense of a challenging semantic trigger-based attack and created a new dataset for the same. PGBD is a scalable and robust post-hoc defense method on which much can be built in the future. The PAV-alignment based loss can possibly be replaced by a distribution distance for better performance on datasets with every uneven distributions. It is easy to extend the activation-space manipulation to other classification problems but we believe the core idea can also be extended to reconstruction problems. We intend to work on such improvements to make PGBD practical in many real-world situations. Finally, it would be interesting to explore the validity of the geometric insights gathered here to other problems such as domain adaptation, feature disentanglement, multi-task learning \etc.

{
    \small
    \bibliographystyle{ieeenat_fullname}
    \bibliography{main}

\begin{thebibliography}{79}
\providecommand{\natexlab}[1]{#1}
\providecommand{\url}[1]{\texttt{#1}}
\expandafter\ifx\csname urlstyle\endcsname\relax
  \providecommand{\doi}[1]{doi: #1}\else
  \providecommand{\doi}{doi: \begingroup \urlstyle{rm}\Url}\fi

\bibitem[Barni et~al.(2019)Barni, Kallas, and Tondi]{Barni2019ANB_signal}
Mauro Barni, Kassem Kallas, and Benedetta Tondi.
\newblock A new backdoor attack in cnns by training set corruption without label poisoning.
\newblock \emph{2019 IEEE International Conference on Image Processing (ICIP)}, 2019.

\bibitem[Biggio et~al.(2011)Biggio, Corona, Fumera, Giacinto, and Roli]{10.1007/978-3-642-21557-5_37}
Battista Biggio, Igino Corona, Giorgio Fumera, Giorgio Giacinto, and Fabio Roli.
\newblock Bagging classifiers for fighting poisoning attacks in adversarial classification tasks.
\newblock In \emph{Multiple Classifier Systems}. Springer Berlin Heidelberg, 2011.

\bibitem[Bontempelli et~al.(2022)Bontempelli, Teso, Tentori, Giunchiglia, and Passerini]{bontempelli2022concept}
Andrea Bontempelli, Stefano Teso, Katya Tentori, Fausto Giunchiglia, and Andrea Passerini.
\newblock Concept-level debugging of part-prototype networks.
\newblock \emph{arXiv preprint arXiv:2205.15769}, 2022.

\bibitem[Borgnia et~al.(2021)Borgnia, Cherepanova, Fowl, Ghiasi, Geiping, Goldblum, Goldstein, and Gupta]{borgnia2021strong}
Eitan Borgnia, Valeriia Cherepanova, Liam Fowl, Amin Ghiasi, Jonas Geiping, Micah Goldblum, Tom Goldstein, and Arjun Gupta.
\newblock Strong data augmentation sanitizes poisoning and backdoor attacks without an accuracy tradeoff.
\newblock In \emph{ICASSP 2021-2021 IEEE International Conference on Acoustics, Speech and Signal Processing (ICASSP)}. IEEE, 2021.

\bibitem[Caron et~al.(2021)Caron, Touvron, Misra, J\'egou, Mairal, Bojanowski, and Joulin]{caron2021emerging}
Mathilde Caron, Hugo Touvron, Ishan Misra, Herv\'e J\'egou, Julien Mairal, Piotr Bojanowski, and Armand Joulin.
\newblock Emerging properties in self-supervised vision transformers.
\newblock In \emph{Proceedings of the International Conference on Computer Vision (ICCV)}, 2021.

\bibitem[Chauhan et~al.(2023)Chauhan, Tiwari, Freyberg, Shenoy, and Dvijotham]{chauhan2023interactive}
Kushal Chauhan, Rishabh Tiwari, Jan Freyberg, Pradeep Shenoy, and Krishnamurthy Dvijotham.
\newblock Interactive concept bottleneck models.
\newblock In \emph{Proceedings of the AAAI Conference on Artificial Intelligence}, pages 5948--5955, 2023.

\bibitem[Chen et~al.(2018)Chen, Carvalho, Baracaldo, Ludwig, Edwards, Lee, Molloy, and Srivastava]{chen2018detecting}
Bryant Chen, Wilka Carvalho, Nathalie Baracaldo, Heiko Ludwig, Benjamin Edwards, Taesung Lee, Ian Molloy, and Biplav Srivastava.
\newblock Detecting backdoor attacks on deep neural networks by activation clustering.
\newblock \emph{arXiv preprint arXiv:1811.03728}, 2018.

\bibitem[Chen et~al.(2021)Chen, Li, Song, and Zhuo]{DBLP:conf/iclr/ChenL0Z21}
Sitan Chen, Xiaoxiao Li, Zhao Song, and Danyang Zhuo.
\newblock On instahide, phase retrieval, and sparse matrix factorization.
\newblock In \emph{9th International Conference on Learning Representations, {ICLR} 2021, Virtual Event, Austria, May 3-7, 2021}. OpenReview.net, 2021.

\bibitem[Chen et~al.(2017)Chen, Liu, Li, Lu, and Song]{Chen2017TargetedBA}
Xinyun Chen, Chang Liu, Bo Li, Kimberly Lu, and Dawn~Xiaodong Song.
\newblock Targeted backdoor attacks on deep learning systems using data poisoning.
\newblock \emph{ArXiv}, abs/1712.05526, 2017.

\bibitem[Chen et~al.(2023)Chen, Guo, Tao, Zhang, and Song]{rl_chen2023bird}
Xuan Chen, Wenbo Guo, Guanhong Tao, Xiangyu Zhang, and Dawn Song.
\newblock Bird: generalizable backdoor detection and removal for deep reinforcement learning.
\newblock \emph{Advances in Neural Information Processing Systems}, 36:\penalty0 40786--40798, 2023.

\bibitem[Cheng et~al.(2021)Cheng, Liu, Ma, and Zhang]{Cheng_Liu_Ma_Zhang_2021}
Siyuan Cheng, Yingqi Liu, Shiqing Ma, and Xiangyu Zhang.
\newblock Deep feature space trojan attack of neural networks by controlled detoxification.
\newblock \emph{Proceedings of the AAAI Conference on Artificial Intelligence}, 35\penalty0 (2), 2021.

\bibitem[Chou et~al.(2020)Chou, Tramer, and Pellegrino]{chou2020sentinet}
Edward Chou, Florian Tramer, and Giancarlo Pellegrino.
\newblock Sentinet: Detecting localized universal attacks against deep learning systems.
\newblock In \emph{2020 IEEE Security and Privacy Workshops (SPW)}. IEEE, 2020.

\bibitem[Cin{\`a} et~al.(2022)Cin{\`a}, Grosse, Demontis, Vascon, Zellinger, Moser, Oprea, Biggio, Pelillo, and Roli]{Cin2022WildPR}
Antonio~Emanuele Cin{\`a}, Kathrin Grosse, Ambra Demontis, Sebastiano Vascon, Werner Zellinger, Bernhard~Alois Moser, Alina Oprea, Battista Biggio, Marcello Pelillo, and Fabio Roli.
\newblock Wild patterns reloaded: A survey of machine learning security against training data poisoning.
\newblock \emph{ACM Computing Surveys}, 55, 2022.

\bibitem[Cohen et~al.(2019)Cohen, Rosenfeld, and Kolter]{pmlr-v97-cohen19c}
Jeremy Cohen, Elan Rosenfeld, and Zico Kolter.
\newblock Certified adversarial robustness via randomized smoothing.
\newblock In \emph{Proceedings of the 36th International Conference on Machine Learning}. PMLR, 2019.

\bibitem[Dong et~al.(2024)Dong, Chen, Zheng, Fu, Zukaib, Cui, and Shen]{dong2024ocie}
Liang Dong, Leiyang Chen, Chengliang Zheng, Zhongwang Fu, Umer Zukaib, Xiaohui Cui, and Zhidong Shen.
\newblock Ocie: Augmenting model interpretability via deconfounded explanation-guided learning.
\newblock \emph{Knowledge-Based Systems}, 302:\penalty0 112390, 2024.

\bibitem[Erak$\iota$n et~al.(2021)Erak$\iota$n, Demir, and Ekenel]{erakiotan2021recognizing}
Mustafa~Ekrem Erak$\iota$n, U{\u{g}}ur Demir, and Haz$\iota$m~Kemal Ekenel.
\newblock On recognizing occluded faces in the wild.
\newblock In \emph{2021 International Conference of the Biometrics Special Interest Group (BIOSIG)}. IEEE, 2021.

\bibitem[Feng et~al.(2023)Feng, Tao, Cheng, Shen, Xu, Liu, Zhang, Ma, and Zhang]{ssl_feng2023detecting}
Shiwei Feng, Guanhong Tao, Siyuan Cheng, Guangyu Shen, Xiangzhe Xu, Yingqi Liu, Kaiyuan Zhang, Shiqing Ma, and Xiangyu Zhang.
\newblock Detecting backdoors in pre-trained encoders.
\newblock In \emph{Proceedings of the IEEE/CVF Conference on Computer Vision and Pattern Recognition}, pages 16352--16362, 2023.

\bibitem[Gao et~al.(2019)Gao, Xu, Wang, Chen, Ranasinghe, and Nepal]{gao2019strip}
Yansong Gao, Change Xu, Derui Wang, Shiping Chen, Damith~C Ranasinghe, and Surya Nepal.
\newblock Strip: A defence against trojan attacks on deep neural networks.
\newblock In \emph{Proceedings of the 35th annual computer security applications conference}, 2019.

\bibitem[Goldblum et~al.(2022)Goldblum, Tsipras, Xie, Chen, Schwarzschild, Song, M{\k{a}}dry, Li, and Goldstein]{goldblum2022dataset}
Micah Goldblum, Dimitris Tsipras, Chulin Xie, Xinyun Chen, Avi Schwarzschild, Dawn Song, Aleksander M{\k{a}}dry, Bo Li, and Tom Goldstein.
\newblock Dataset security for machine learning: Data poisoning, backdoor attacks, and defenses.
\newblock \emph{IEEE Transactions on Pattern Analysis and Machine Intelligence}, 45\penalty0 (2):\penalty0 1563--1580, 2022.

\bibitem[Gu et~al.(2019)Gu, Liu, Dolan-Gavitt, and Garg]{8685687}
Tianyu Gu, Kang Liu, Brendan Dolan-Gavitt, and Siddharth Garg.
\newblock Badnets: Evaluating backdooring attacks on deep neural networks.
\newblock \emph{IEEE Access}, 7, 2019.

\bibitem[Guo et~al.(2020)Guo, Wang, Xu, Xing, Du, and Song]{TABOR}
Wenbo Guo, Lun Wang, Yan Xu, Xinyu Xing, Min Du, and Dawn Song.
\newblock Towards inspecting and eliminating trojan backdoors in deep neural networks.
\newblock In \emph{2020 IEEE International Conference on Data Mining (ICDM)}, pages 162--171, 2020.

\bibitem[Gupta et~al.(2023)Gupta, Saini, and Narayanan]{Gupta2023ConceptDL}
Avani Gupta, Saurabh Saini, and P~J Narayanan.
\newblock Concept distillation: Leveraging human-centered explanations for model improvement.
\newblock In \emph{NeurIPS}, 2023.

\bibitem[Hayase et~al.(2021)Hayase, Kong, Somani, and Oh]{pmlr-v139-hayase21a}
Jonathan Hayase, Weihao Kong, Raghav Somani, and Sewoong Oh.
\newblock Spectre: defending against backdoor attacks using robust statistics.
\newblock In \emph{Proceedings of the 38th International Conference on Machine Learning}, 2021.

\bibitem[He et~al.(2020)He, Xue, Wang, and Liu]{10.1145/3393527.3393567}
Can He, Mingfu Xue, Jian Wang, and Weiqiang Liu.
\newblock Embedding backdoors as the facial features: Invisible backdoor attacks against face recognition systems.
\newblock In \emph{Proceedings of the ACM Turing Celebration Conference - China}. Association for Computing Machinery, 2020.

\bibitem[He et~al.(2016)He, Zhang, Ren, and Sun]{he2016identity}
Kaiming He, Xiangyu Zhang, Shaoqing Ren, and Jian Sun.
\newblock Identity mappings in deep residual networks.
\newblock In \emph{Computer Vision--ECCV 2016: 14th European Conference, Amsterdam, The Netherlands, October 11--14, 2016, Proceedings, Part IV 14}. Springer, 2016.

\bibitem[He et~al.(2021)He, Meng, Chen, He, and Hu]{He2021DeepObliviateAP}
Yingzhe He, Guozhu Meng, Kai Chen, Jinwen He, and Xingbo Hu.
\newblock Deepobliviate: A powerful charm for erasing data residual memory in deep neural networks.
\newblock \emph{ArXiv}, abs/2105.06209, 2021.

\bibitem[Hu et~al.(2021)Hu, Lin, Cogswell, Yao, Jha, and Chen]{hu2021trigger}
Xiaoling Hu, Xiao Lin, Michael Cogswell, Yi Yao, Susmit Jha, and Chao Chen.
\newblock Trigger hunting with a topological prior for trojan detection.
\newblock \emph{arXiv preprint arXiv:2110.08335}, 2021.

\bibitem[Jia et~al.(2021)Jia, Cao, and Gong]{certified_jia2021intrinsic}
Jinyuan Jia, Xiaoyu Cao, and Neil~Zhenqiang Gong.
\newblock Intrinsic certified robustness of bagging against data poisoning attacks.
\newblock In \emph{Proceedings of the AAAI conference on artificial intelligence}, pages 7961--7969, 2021.

\bibitem[Keswani et~al.(2022)Keswani, Ramakrishnan, Reddy, and Balasubramanian]{keswani2022proto2proto}
Monish Keswani, Sriranjani Ramakrishnan, Nishant Reddy, and Vineeth~N Balasubramanian.
\newblock Proto2proto: Can you recognize the car, the way i do?
\newblock In \emph{Proceedings of the IEEE/CVF Conference on Computer Vision and Pattern Recognition}, 2022.

\bibitem[Kim et~al.(2018)Kim, Wattenberg, Gilmer, Cai, Wexler, Viegas, and Sayres]{tcav}
Been Kim, Martin Wattenberg, Justin Gilmer, Carrie~Jun Cai, James Wexler, Fernanda Viegas, and Rory~Abbott Sayres.
\newblock Interpretability beyond feature attribution: Quantitative testing with concept activation vectors (tcav).
\newblock 2018.

\bibitem[Koh et~al.(2020)Koh, Nguyen, Tang, Mussmann, Pierson, Kim, and Liang]{koh2020concept}
Pang~Wei Koh, Thao Nguyen, Yew~Siang Tang, Stephen Mussmann, Emma Pierson, Been Kim, and Percy Liang.
\newblock Concept bottleneck models.
\newblock In \emph{International conference on machine learning}, pages 5338--5348. PMLR, 2020.

\bibitem[Kolouri et~al.(2020)Kolouri, Saha, Pirsiavash, and Hoffmann]{kolouri2020universal}
Soheil Kolouri, Aniruddha Saha, Hamed Pirsiavash, and Heiko Hoffmann.
\newblock Universal litmus patterns: Revealing backdoor attacks in cnns.
\newblock In \emph{Proceedings of the IEEE/CVF Conference on Computer Vision and Pattern Recognition}, pages 301--310, 2020.

\bibitem[Krizhevsky(2009)]{krizhevsky2009learning}
Alex Krizhevsky.
\newblock Learning multiple layers of features from tiny images.
\newblock 2009.

\bibitem[Le and Yang(2015)]{le2015tiny}
Ya Le and Xuan Yang.
\newblock Tiny imagenet visual recognition challenge.
\newblock \emph{CS 231N}, 7\penalty0 (7):\penalty0 3, 2015.

\bibitem[Li et~al.(2023{\natexlab{a}})Li, Pang, Xi, Du, Ji, Yao, and Wang]{ssl_li2023embarrassingly}
Changjiang Li, Ren Pang, Zhaohan Xi, Tianyu Du, Shouling Ji, Yuan Yao, and Ting Wang.
\newblock An embarrassingly simple backdoor attack on self-supervised learning.
\newblock In \emph{Proceedings of the IEEE/CVF International Conference on Computer Vision}, pages 4367--4378, 2023{\natexlab{a}}.

\bibitem[Li et~al.(2021{\natexlab{a}})Li, Koren, Lyu, Lyu, Li, and Ma]{Li2021NeuralAD}
Yige Li, Nodens Koren, L. Lyu, Xixiang Lyu, Bo Li, and Xingjun Ma.
\newblock Neural attention distillation: Erasing backdoor triggers from deep neural networks.
\newblock \emph{ArXiv}, abs/2101.05930, 2021{\natexlab{a}}.

\bibitem[Li et~al.(2021{\natexlab{b}})Li, Li, Wu, Li, He, and Lyu]{ssba_li2021invisible}
Yuezun Li, Yiming Li, Baoyuan Wu, Longkang Li, Ran He, and Siwei Lyu.
\newblock Invisible backdoor attack with sample-specific triggers.
\newblock In \emph{Proceedings of the IEEE/CVF international conference on computer vision}, pages 16463--16472, 2021{\natexlab{b}}.

\bibitem[Li et~al.(2021{\natexlab{c}})Li, Lyu, Koren, Lyu, Li, and Ma]{li2021anti}
Yige Li, Xixiang Lyu, Nodens Koren, Lingjuan Lyu, Bo Li, and Xingjun Ma.
\newblock Anti-backdoor learning: Training clean models on poisoned data.
\newblock In \emph{NeurIPS}, 2021{\natexlab{c}}.

\bibitem[Li et~al.(2023{\natexlab{b}})Li, Lyu, Ma, Koren, Lyu, Li, and Jiang]{pmlr-v202-li23v}
Yige Li, Xixiang Lyu, Xingjun Ma, Nodens Koren, Lingjuan Lyu, Bo Li, and Yu-Gang Jiang.
\newblock Reconstructive neuron pruning for backdoor defense.
\newblock In \emph{Proceedings of the 40th International Conference on Machine Learning}, pages 19837--19854. PMLR, 2023{\natexlab{b}}.

\bibitem[Liu et~al.(2019)Liu, Lee, Tao, Ma, Aafer, and Zhang]{10.1145/3319535.3363216}
Yingqi Liu, Wen-Chuan Lee, Guanhong Tao, Shiqing Ma, Yousra Aafer, and Xiangyu Zhang.
\newblock Abs: Scanning neural networks for back-doors by artificial brain stimulation.
\newblock In \emph{Proceedings of the 2019 ACM SIGSAC Conference on Computer and Communications Security}. Association for Computing Machinery, 2019.

\bibitem[Liu et~al.(2020)Liu, Ma, Bailey, and Lu]{Liu2020Refool}
Yunfei Liu, Xingjun Ma, James Bailey, and Feng Lu.
\newblock Reflection backdoor: A natural backdoor attack on deep neural networks.
\newblock In \emph{ECCV}, 2020.

\bibitem[Liu et~al.(2022)Liu, Fan, Chen, Liu, Ma, Wang, and Ma]{backdoor_unlearning}
Yang Liu, Mingyuan Fan, Cen Chen, Ximeng Liu, Zhuo Ma, Li Wang, and Jianfeng Ma.
\newblock Backdoor defense with machine unlearning.
\newblock In \emph{IEEE INFOCOM 2022 - IEEE Conference on Computer Communications}, pages 280--289, 2022.

\bibitem[Min et~al.(2023)Min, Qin, Shen, and Cheng]{min2023towards_fst}
Rui Min, Zeyu Qin, Li Shen, and Minhao Cheng.
\newblock Towards stable backdoor purification through feature shift tuning.
\newblock \emph{Advances in Neural Information Processing Systems}, 36:\penalty0 75286--75306, 2023.

\bibitem[Nguyen and Tran(2020)]{iab_nguyen2020input}
Tuan~Anh Nguyen and Anh Tran.
\newblock Input-aware dynamic backdoor attack.
\newblock \emph{Advances in Neural Information Processing Systems}, 33:\penalty0 3454--3464, 2020.

\bibitem[Nguyen and Tran(2021)]{nguyen2021wanet}
Tuan~Anh Nguyen and Anh~Tuan Tran.
\newblock Wanet - imperceptible warping-based backdoor attack.
\newblock In \emph{International Conference on Learning Representations}, 2021.

\bibitem[Oquab et~al.(2023)Oquab, Darcet, Moutakanni, Vo, Szafraniec, Khalidov, Fernandez, Haziza, Massa, El-Nouby, Howes, Huang, Xu, Sharma, Li, Galuba, Rabbat, Assran, Ballas, Synnaeve, Misra, Jegou, Mairal, Labatut, Joulin, and Bojanowski]{oquab2023dinov2}
Maxime Oquab, Timothée Darcet, Theo Moutakanni, Huy~V. Vo, Marc Szafraniec, Vasil Khalidov, Pierre Fernandez, Daniel Haziza, Francisco Massa, Alaaeldin El-Nouby, Russell Howes, Po-Yao Huang, Hu Xu, Vasu Sharma, Shang-Wen Li, Wojciech Galuba, Mike Rabbat, Mido Assran, Nicolas Ballas, Gabriel Synnaeve, Ishan Misra, Herve Jegou, Julien Mairal, Patrick Labatut, Armand Joulin, and Piotr Bojanowski.
\newblock Dinov2: Learning robust visual features without supervision, 2023.

\bibitem[Paszke et~al.(2017)Paszke, Gross, Chintala, Chanan, Yang, DeVito, Lin, Desmaison, Antiga, and Lerer]{pytorch}
Adam Paszke, Sam Gross, Soumith Chintala, Gregory Chanan, Edward Yang, Zachary DeVito, Zeming Lin, Alban Desmaison, Luca Antiga, and Adam Lerer.
\newblock Automatic differentiation in pytorch.
\newblock 2017.

\bibitem[Paudice et~al.(2019)Paudice, Mu{\~{n}}oz-Gonz{\'a}lez, and Lupu]{10.1007/978-3-030-13453-2_1}
Andrea Paudice, Luis Mu{\~{n}}oz-Gonz{\'a}lez, and Emil~C. Lupu.
\newblock Label sanitization against label flipping poisoning attacks.
\newblock In \emph{ECML PKDD 2018 Workshops}. Springer International Publishing, 2019.

\bibitem[Saha et~al.(2022)Saha, Tejankar, Koohpayegani, and Pirsiavash]{ssl_saha2022backdoor}
Aniruddha Saha, Ajinkya Tejankar, Soroush~Abbasi Koohpayegani, and Hamed Pirsiavash.
\newblock Backdoor attacks on self-supervised learning.
\newblock In \emph{Proceedings of the IEEE/CVF Conference on Computer Vision and Pattern Recognition}, pages 13337--13346, 2022.

\bibitem[Salem et~al.(2022)Salem, Wen, Backes, Ma, and Zhang]{dynamicba_salem2022dynamic}
Ahmed Salem, Rui Wen, Michael Backes, Shiqing Ma, and Yang Zhang.
\newblock Dynamic backdoor attacks against machine learning models.
\newblock In \emph{2022 IEEE 7th European Symposium on Security and Privacy (EuroS\&P)}, pages 703--718. IEEE, 2022.

\bibitem[Sarkar et~al.(2022)Sarkar, Benkraouda, Krishnan, Gamil, and Maniatakos]{Sarkar2022FaceHackAF}
Esha Sarkar, Hadjer Benkraouda, G~Muthu Krishnan, Homer Gamil, and Michail Maniatakos.
\newblock Facehack: Attacking facial recognition systems using malicious facial characteristics.
\newblock \emph{IEEE Transactions on Biometrics, Behavior, and Identity Science}, 4, 2022.

\bibitem[Selvaraju et~al.(2017)Selvaraju, Cogswell, Das, Vedantam, Parikh, and Batra]{gradcam}
Ramprasaath~R. Selvaraju, Michael Cogswell, Abhishek Das, Ramakrishna Vedantam, Devi Parikh, and Dhruv Batra.
\newblock Grad-cam: Visual explanations from deep networks via gradient-based localization.
\newblock In \emph{2017 IEEE International Conference on Computer Vision (ICCV)}, pages 618--626, 2017.

\bibitem[Song et~al.(2023)Song, Shyn, and Kim]{song2023img2tab}
Youngjae Song, Sung~Kuk Shyn, and Kwang-su Kim.
\newblock Img2tab: Automatic class relevant concept discovery from stylegan features for explainable image classification.
\newblock \emph{arXiv preprint arXiv:2301.06324}, 2023.

\bibitem[Souri et~al.(2022)Souri, Fowl, Chellappa, Goldblum, and Goldstein]{souri2022sleeper}
Hossein Souri, Liam Fowl, Rama Chellappa, Micah Goldblum, and Tom Goldstein.
\newblock Sleeper agent: Scalable hidden trigger backdoors for neural networks trained from scratch.
\newblock \emph{Advances in Neural Information Processing Systems}, 35, 2022.

\bibitem[Stallkamp et~al.(2012)Stallkamp, Schlipsing, Salmen, and Igel]{STALLKAMP2012323}
J. Stallkamp, M. Schlipsing, J. Salmen, and C. Igel.
\newblock Man vs. computer: Benchmarking machine learning algorithms for traffic sign recognition.
\newblock \emph{Neural Networks}, 32, 2012.

\bibitem[Taheri et~al.(2020)Taheri, Javidan, Shojafar, Pooranian, Miri, and Conti]{Taheri@2020}
Rahim Taheri, Reza Javidan, Mohammad Shojafar, Zahra Pooranian, Ali Miri, and Mauro Conti.
\newblock On defending against label flipping attacks on malware detection systems.
\newblock \emph{Neural Computing and Applications}, 32, 2020.

\bibitem[Tang et~al.(2020)Tang, Du, Liu, Yang, and Hu]{10.1145/3394486.3403064}
Ruixiang Tang, Mengnan Du, Ninghao Liu, Fan Yang, and Xia Hu.
\newblock An embarrassingly simple approach for trojan attack in deep neural networks.
\newblock In \emph{Proceedings of the 26th ACM SIGKDD International Conference on Knowledge Discovery \& Data Mining}. Association for Computing Machinery, 2020.

\bibitem[Tao et~al.(2022)Tao, Shen, Liu, An, Xu, Ma, and Zhang]{Tao2022BetterTI}
Guanhong Tao, Guangyu Shen, Yingqi Liu, Shengwei An, Qiuling Xu, Shiqing Ma, and X. Zhang.
\newblock Better trigger inversion optimization in backdoor scanning.
\newblock \emph{2022 IEEE/CVF Conference on Computer Vision and Pattern Recognition (CVPR)}, 2022.

\bibitem[Tejankar et~al.(2023)Tejankar, Sanjabi, Wang, Wang, Firooz, Pirsiavash, and Tan]{ssl_tejankar2023defending}
Ajinkya Tejankar, Maziar Sanjabi, Qifan Wang, Sinong Wang, Hamed Firooz, Hamed Pirsiavash, and Liang Tan.
\newblock Defending against patch-based backdoor attacks on self-supervised learning.
\newblock In \emph{Proceedings of the IEEE/CVF conference on computer vision and pattern recognition}, pages 12239--12249, 2023.

\bibitem[Tran et~al.(2018)Tran, Li, and Madry]{Tran2018SpectralSI}
Brandon Tran, Jerry Li, and Aleksander Madry.
\newblock Spectral signatures in backdoor attacks.
\newblock In \emph{Neural Information Processing Systems}, 2018.

\bibitem[Tschernezki et~al.(2022)Tschernezki, Laina, Larlus, and Vedaldi]{tschernezki22neural}
Vadim Tschernezki, Iro Laina, Diane Larlus, and Andrea Vedaldi.
\newblock Neural feature fusion fields: {3D} distillation of self-supervised {2D} image representations.
\newblock In \emph{Proceedings of the International Conference on {3D} Vision (3DV)}, 2022.

\bibitem[van~der Maaten and Hinton(2008)]{tsne}
Laurens van~der Maaten and Geoffrey Hinton.
\newblock Visualizing data using t-sne.
\newblock \emph{Journal of Machine Learning Research}, 9\penalty0 (86):\penalty0 2579--2605, 2008.

\bibitem[Wang et~al.(2019)Wang, Yao, Shan, Li, Viswanath, Zheng, and Zhao]{Wang2019NeuralCI}
Bolun Wang, Yuanshun Yao, Shawn Shan, Huiying Li, Bimal Viswanath, Haitao Zheng, and Ben~Y. Zhao.
\newblock Neural cleanse: Identifying and mitigating backdoor attacks in neural networks.
\newblock \emph{2019 IEEE Symposium on Security and Privacy (SP)}, 2019.

\bibitem[Wang et~al.(2021)Wang, Javed, Wu, Guo, Xing, and Song]{rl_wang2021backdoorl}
Lun Wang, Zaynah Javed, Xian Wu, Wenbo Guo, Xinyu Xing, and Dawn Song.
\newblock Backdoorl: Backdoor attack against competitive reinforcement learning.
\newblock \emph{arXiv preprint arXiv:2105.00579}, 2021.

\bibitem[Wang et~al.(2022)Wang, Levine, and Feizi]{pmlr-v162-wang22m}
Wenxiao Wang, Alexander~J Levine, and Soheil Feizi.
\newblock Improved certified defenses against data poisoning with ({D}eterministic) finite aggregation.
\newblock In \emph{Proceedings of the 39th International Conference on Machine Learning}. PMLR, 2022.

\bibitem[Wenger et~al.(2021)Wenger, Passananti, Bhagoji, Yao, Zheng, and Zhao]{9577800}
E. Wenger, J. Passananti, A. Bhagoji, Y. Yao, H. Zheng, and B.~Y. Zhao.
\newblock Backdoor attacks against deep learning systems in the physical world.
\newblock In \emph{2021 IEEE/CVF Conference on Computer Vision and Pattern Recognition (CVPR)}. IEEE Computer Society, 2021.

\bibitem[Wu et~al.(2022)Wu, Chen, Zhang, Zhu, Wei, Yuan, and Shen]{wu2022backdoorbench}
Baoyuan Wu, Hongrui Chen, Mingda Zhang, Zihao Zhu, Shaokui Wei, Danni Yuan, and Chao Shen.
\newblock Backdoorbench: A comprehensive benchmark of backdoor learning.
\newblock In \emph{Thirty-sixth Conference on Neural Information Processing Systems Datasets and Benchmarks Track}, 2022.

\bibitem[Wu and Wang(2021)]{ANP_NEURIPS2021_8cbe9ce2}
Dongxian Wu and Yisen Wang.
\newblock Adversarial neuron pruning purifies backdoored deep models.
\newblock In \emph{Advances in Neural Information Processing Systems}, pages 16913--16925. Curran Associates, Inc., 2021.

\bibitem[Xia et~al.(2022)Xia, Niu, Li, and Li]{xia2022enhancing_mmdr}
Pengfei Xia, Hongjing Niu, Ziqiang Li, and Bin Li.
\newblock Enhancing backdoor attacks with multi-level mmd regularization.
\newblock \emph{IEEE Transactions on Dependable and Secure Computing}, 2022.

\bibitem[Xiang et~al.(2020)Xiang, Miller, and Kesidis]{xiang2020detection}
Zhen Xiang, David~J Miller, and George Kesidis.
\newblock Detection of backdoors in trained classifiers without access to the training set.
\newblock \emph{IEEE Transactions on Neural Networks and Learning Systems}, 33\penalty0 (3), 2020.

\bibitem[Xiang et~al.(2021)Xiang, Miller, and Kesidis]{10.1016/j.cose.2021.102280}
Zhen Xiang, David~J. Miller, and George Kesidis.
\newblock Reverse engineering imperceptible backdoor attacks on deep neural networks for detection and training set cleansing.
\newblock \emph{Comput. Secur.}, 106\penalty0 (C), 2021.

\bibitem[Xiang et~al.(2023)Xiang, Xiong, and Li]{certified_xiang2023cbd}
Zhen Xiang, Zidi Xiong, and Bo Li.
\newblock Cbd: A certified backdoor detector based on local dominant probability.
\newblock \emph{Advances in Neural Information Processing Systems}, 36:\penalty0 4937--4951, 2023.

\bibitem[Xu et~al.(2021)Xu, Wang, Li, Borisov, Gunter, and Li]{xu2021detecting}
Xiaojun Xu, Qi Wang, Huichen Li, Nikita Borisov, Carl~A Gunter, and Bo Li.
\newblock Detecting ai trojans using meta neural analysis.
\newblock In \emph{2021 IEEE Symposium on Security and Privacy (SP)}. IEEE, 2021.

\bibitem[Yao et~al.(2019)Yao, Li, Zheng, and Zhao]{10.1145/3319535.3354209}
Yuanshun Yao, Huiying Li, Haitao Zheng, and Ben~Y. Zhao.
\newblock Latent backdoor attacks on deep neural networks.
\newblock In \emph{Proceedings of the 2019 ACM SIGSAC Conference on Computer and Communications Security}. Association for Computing Machinery, 2019.

\bibitem[Yue et~al.(2023)Yue, Xia, Ling, Hu, Wang, Wei, and Chen]{Yue2023ModelContrastiveLF}
Zhihao Yue, Jun Xia, Zhiwei Ling, Ming Hu, Ting Wang, Xian Wei, and Mingsong Chen.
\newblock Model-contrastive learning for backdoor elimination.
\newblock In \emph{Proceedings of the 31st ACM International Conference on Multimedia}, pages 8869--8880, 2023.

\bibitem[Zeng et~al.(2021)Zeng, Chen, Park, Mao, Jin, and Jia]{Zeng2021AdversarialUO}
Yi Zeng, Si Chen, Won Park, Zhuoqing Mao, Ming Jin, and Ruoxi Jia.
\newblock Adversarial unlearning of backdoors via implicit hypergradient.
\newblock 2021.

\bibitem[Zeng et~al.(2023)Zeng, Shi, Jin, Kang, Lyu, Hsieh, and Jia]{certified_zeng2023towards}
Yi Zeng, Zhouxing Shi, Ming Jin, Feiyang Kang, Lingjuan Lyu, Cho-Jui Hsieh, and Ruoxi Jia.
\newblock Towards robustness certification against universal perturbations.
\newblock In \emph{International Conference on Learning Representation}. ICLR, 2023.

\bibitem[Zhu et~al.(2020)Zhu, Ning, Wang, Xin, and Wu]{zhu2020gangsweep}
Liuwan Zhu, Rui Ning, Cong Wang, Chunsheng Xin, and Hongyi Wu.
\newblock Gangsweep: Sweep out neural backdoors by gan.
\newblock In \emph{Proceedings of the 28th ACM International Conference on Multimedia}, 2020.

\bibitem[Zhu et~al.(2023)Zhu, Wei, Shen, Fan, and Wu]{ft_sam_Zhu_2023_ICCV}
Mingli Zhu, Shaokui Wei, Li Shen, Yanbo Fan, and Baoyuan Wu.
\newblock Enhancing fine-tuning based backdoor defense with sharpness-aware minimization.
\newblock In \emph{Proceedings of the IEEE/CVF International Conference on Computer Vision (ICCV)}, pages 4466--4477, 2023.

\end{thebibliography}
}
\clearpage
\appendix


\section*{Appendix}
We present additional results, more details on the method, and visualizations to display the robustness, scalability, and configurability of PGBD. The contents are roughly as given below.
\begin{enumerate}
    \item Additional Details on the Method (\Cref{sec:supple_method})
    \item Results on Variations of PGBD (\Cref{sec:supple_variations})
    \item Implementation Details (\Cref{sec:supple_implementation})
    \item Semantic Attack Dataset: Details (\Cref{sec:supple_datasets})
    \item Quantitative Ablation Results (\Cref{sec:ablations_supple})
    \item Additional Result Visualizations (\Cref{sec:supple_visualizations})
\end{enumerate}

\section{Additional Details on the Method }
\label{sec:supple_method}
\subsection{PAV updates}
We periodically update the prototypes ($P$) and the PAVs ($V$) while performing defense. This is required especially when the change in model activation space is significant, which is the case in most of our runs. Specifically, an update implies recalculating the prototypes using new activations of the model on $D_s$. These recalculated prototypes $P_new$ are not used directly, but are weighted with the previous prototypes $P_old$ to perform a \emph{slow update} as shown below:
\begin{equation} \label{eqn:protoSlowUpdate}
    P = (1 - \alpha)P_{old} + \alpha P_{new} 
\end{equation}
Therefore, $\alpha$ controls the extent of update we want to factor in when updating the PAVs and prototypes. We use $\alpha=0.75$ throughout the paper (unless mentioned otherwise), and show effects on ablating in \Cref{fig:abl_alpha}.

\subsection{Algorithm}
The step-by-step algorithm of our overall method discussed in the main paper is given below. 
\begin{center}
    \RestyleAlgo{ruled}
    \begin{algorithm}
    \caption{Prototype Guided Backdoor Defense}
    \label{alg:PBD}
    \footnotesize
    \SetCommentSty{textit}
    \KwIn{Backdoor Model:$M_{B}$; Clean dataset:$D_{s}$; Params. :$ \lambda, \alpha$; Dino: $\phi$}
    \KwOut{Clean Model: $M_{C}$}
    $t \gets {NeuralCleanse}(M_{B}, D_{s})$   \tcp*[r]{Get target label $t$ using prev works.}
    \textcolor{cyan}{$p_t \gets {TriggerSyn}(M_B,t)$}  \tcp*[r]{[Optional] Get trigger for target class \cite{Tao2022BetterTI} }
    \textcolor{cyan}{$D'_s \gets {Poison}(D_s,p_t)$}     \tcp*[r]{[Optional] Simulate poisoned dataset}
    \tcp{Optional steps using $V^s$ shown in \textcolor{cyan}{blue} separated by $`|'$}
    \For{$e \gets 0$ \KwTo {num of epochs}}
    {
        \tcp{Estimate PAV}
        \If{$e \% 5 == 0$}
        {
            \tcp{Train Mapping Module}
            $M_B(D_s) \gets Encoder(\phi(D_s))$ \tcp*[r]{Teacher to student}
            $\phi(D_s) \gets Decoder(M_B(D_s))$ \tcp*[r]{Student to teacher}
            \For{$c \in \{0,n\}$}
            {
                \tcp{Compute clean and poisoned class prototypes in student}
                \textcolor{black}{$P_c \gets {Kmeans} (D_s)$} $\;|\;$ \textcolor{cyan}{$P'_c \gets {Kmeans} (D'_s)$} \\
                $P_c^e \gets (1 - \alpha)*P_{c}^{e-1} + \alpha*P_{c}^{e}  \quad \forall e>0$ \tcp*[r]{\Cref{eqn:protoSlowUpdate}}
                \tcp{Estimate PAVs in teacher and remap to student}
                \textcolor{black}{$\phi(P_c) \gets  Decoder(P_c)$} $\;\;\;\;|\;$ \textcolor{cyan}{$\phi(P'_c) \gets Decoder(P'_c)$}  \\
                \textcolor{black}{$\phi(V^{p}) \gets \phi(P_{t}) - \phi(P_{c})$} $\;|\;$
                \textcolor{cyan}{$\phi(V^{s}) \gets  \phi(P'_{c}) - \phi(P_{c})$} \\
                \textcolor{black}{$V^{p} \gets Encoder(\phi(V^{p}))$}  $\;\;\,|\;$
                \textcolor{cyan}{$V^{s} \gets  Encoder(\phi(V^{s}))$} \\
            }
        }
        \tcp{Sanitize Model}
        \For{$(x,y) \in D_{s}$}
        {    
            $L_p \gets mse(f(x),P_{c=y})$ \tcp*[r]{$y$ is GT label of $x$ Eqn2 (main paper)} 
            $L_s \gets cos\theta(\nabla L_p, V^p \;|\; \textcolor{cyan}{V^s})$  \tcp*[r]{cosine similarity (Eqn 3, main paper)}
            $L \gets L_{o} + \lambda * L_{s}$ \\ 
            Backpropagate $L$ \;
        }
    }
    \end{algorithm}
\end{center}
\vspace*{-\baselineskip}

\begin{table*}[t!]
\centering
\begin{tabular}{c|l||cc||ccc|ccc|ccc}
\toprule[1.2pt]
  \multicolumn{2}{c||}{\bf Method \faCaretRight} &
  \multicolumn{2}{c||}{\bf Baseline} &
  \multicolumn{3}{c|}{\bf NT-PGBD} &
  \multicolumn{3}{c|}{\bf ST-PGBD} &
  \multicolumn{3}{c}{\bf PGBD} \\
\hline 
 & Attack & CA  & ASR  & CA $\uparrow$ & ASR $\downarrow$ & $\Gamma$ $\uparrow$ & CA $\uparrow$  & ASR $\downarrow$ &  $\Gamma$ $\uparrow$ & CA $\uparrow$ & ASR $\downarrow$ & $\Gamma$ $\uparrow$ \\
\midrule[0.5pt]
\multirow{5}{*}{\STAB{\rotatebox[origin=c]{90}{CIFAR10}}}
& Badnet & 92.34 & 88.93 & 89.44 & 1.6 & \underline{0.98} & 91.22 & 1.70 & \underline{0.98} & 90.66 & 0.82 & \textbf{0.99} \\
& Trojan & 93 & 100 & 73.2 & 19.54 & 0.80 & 76.34 & 2.68 & \underline{0.98} & 83.60 & 6.76 & \textbf{0.92} \\
& Blended & 93 & 92.9 & 83.48 & 13.3 & 0.88 & 87.51 & 2.58 & \textbf{0.96} & 86.11 & 4.87 & \underline{0.94} \\
& Signal & 92.9 & 89.04 & 84.01 & 0.49 & \underline{0.95} & 89.94 & 7.51 & 0.94 & 87.18 & 0.31 & \textbf{0.97} \\
& Wanet & 89.98 & 97.58 & 90.65 & 0.26 & \textbf{1.00} & 83.61 & 0.61 & 0.96 & 88.54 & 2.36 & \underline{0.98} \\ 
& IAB & 90.49 & 91.01 & 88.07 & 3.17 & \underline{0.97} & 92.22 & 14.29 & 0.92 & 89.43 & 2.68 & \textbf{0.98} \\

\hline 
\multirow{3}{*}{\STAB{\rotatebox[origin=c]{90}{ROF}}}
& Sunglass & 93.33 & 86.19 & 63.33 & 1.67 & \underline{0.83} & 59.07 & 0.85 & 0.81 & 71.67 & 4.97 & \textbf{0.86} \\
& Tattoo & 78.84 & 72.1 & 74.33 & 3.82 & \underline{0.94} & 48.86 & 1.3 & 0.80 & 86.53 & 2.40 & \textbf{0.98} \\
& Mask & 69.23 & 96.66 & 53.97 & 0.414 & \underline{0.89} & 76.92 & 56.67 & 0.71 & 63.46 & 3.33 & \textbf{0.94} \\
\hline 
\multirow{4}{*}{\STAB{\rotatebox[origin=c]{90}{CIFAR100}}} 
& Badnet & 67.32 & 86.98 & 59.07 & 0.85 & 0.93 & 62.80 & 0.03 & \underline{0.97} & 64.29 & 0.01 & \textbf{0.98} \\
& Trojan & 70.02 & 100.00 & 48.86 & 1.3 & 0.84 & 54.53 & 0.04 & \underline{0.89} & 62.50 & 0.02 & \textbf{0.95} \\
& Blended & 69.01 & 99.48 & 53.97 & 0.414 & \underline{0.89} & 42.89 & 2.03 & 0.80 & 61.44 & 0.00 & \textbf{0.95} \\
& Wanet & 63.84 & 91.47 & 57.51 & 0.58 & \underline{0.95} & 57.04 & 0.90 & 0.94 & 62.34 & 0.66 & \textbf{0.98}\\
\hline 
\multirow{4}{*}{\STAB{\rotatebox[origin=c]{90}{GTSRB}}} 
& Badnet & 96.61 & 83.86 & 96.83 & 0 & \textbf{1.00} & 96.79 & 0.00 & \textbf{1.00} & 97.26 & 0.00 & \textbf{1.00} \\
& Trojan & 98.17 & 100.00 & 93.72 & 0 & \underline{0.98} & 97.18 & 0.06 & \textbf{0.99} & 96.50 & 0.11 & \textbf{0.99} \\
& Blended & 98.66 & 96.33 & 93.85 & 0 & \textbf{0.98} & 89.94 & 3.67 & \underline{0.94} & 86.20 & 0.72 & 0.93 \\
& Wanet & 98.04 & 82.14 & 98.39 & 0.1 & \textbf{1.00} & 97.35 & 0.08 & \textbf{1.00} & 97.14 & 0.33 & \underline{0.99}\\

\bottomrule[1.2pt]
\end{tabular}

\caption{Quantitative comparison between all three variations of PGBD for four benchmarks (CIFAR10, ROF, CIFAR100, GTSRB). Best and second best values are in \textbf{bold} and \underline{underline}, respectively. We observe that PGBD achieves the overall best results, but NT-PGBD is on par. ST-PGBD struggles naturally when the trigger synthesis is bad.}
\label{tab:abl_variations}
\end{table*}
\subsection{Large-Model Mapping}
\begin{figure*}[h]
    \centering
    \includegraphics[width=\linewidth]{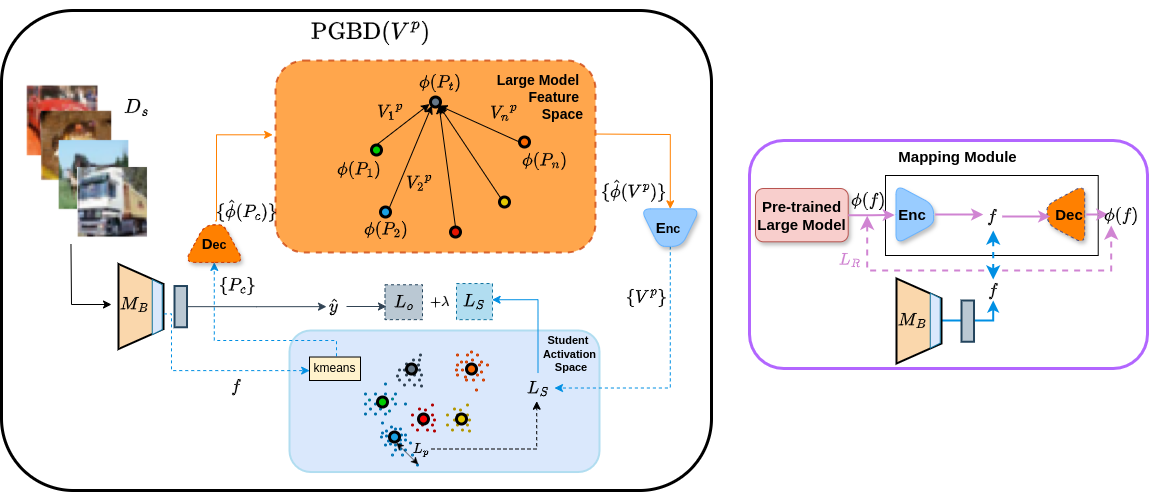}
    \caption{{\bf [Right]} Block diagram for PGBD when using large-model mapping. {\bf [Left]} training procedure of the mapping module using the standard reconstructive loss $L_R$. In the mapping module, the encoder maps from teacher feature $\phi(f)$ to student activation $f$, and the decoder maps from backdoored student $M_B$ activations $f$ to corresponding Large model features $\phi(f)$.}
    \label{fig:distill_block}
\end{figure*}

We take a closer look at the large model mapping process in this section. If using mapping, the first step to perform is training the mapping module. We use a minimal($\approx$ 150 smaples) subset of the available clean data to train the mapping module. As seen in \Cref{fig:distill_block}, the mapping module is a fully convolutional autoencoder, with the bottleneck dimensions resembling student model activations and the input/output layer dimensions matching that of the teacher model. Additionally, the mapping module is lightweight (consisting of only two convolutional layers in total) and linear. We train the overall mapping module for 5 epochs and use the decoder and encoder parts of it seperately as shown in \Cref{fig:distill_block}. Specifically, we use the decoder half to lift student prototypes (${P}$) to the teacher space (where they are represented as $\phi(P)$). We compute the PAVs in the teacher space ($\phi(V)$), where we expect the prototypes to be less cluttered, and map them back to the student space using the encoder half to get the final PAVs($V$). It is important to note that we only perform mapping of the prototypes, and not all the samples in $D_s$. While we believe this is valid given that the mapping module is linear in nature, we verify the same in \Cref{subsec:def_abl}.
\section{Results on Variations of PGBD}
\label{sec:supple_variations}

\subsection{NT-PGBD}
The {\em no target} variation of our method works in the general model sanitization setting where defense methods do not require target label $t$. We achieve this by sanitizing the model by sequentially cycling over all the target labels $t$. We showed in \Cref{tab:variations_mean} that NT-PGBD works on par with the base PGBD but at the cost of greater compute time. Specifically, we chose cycling time as $max(round(\text{num\_epochs}/ N), 1)$, where num\_epochs = 35, and $N$ is the number of classes in the dataset. This increases compute time for datasets with many classes since the total number of epochs is $N*\text{cycle\_time}$. We provide dataset-wise results with cycling times decided based on the aforementioned rules in \Cref{tab:abl_variations}. 

\subsection{ST-PGBD}
This variation of our method works in the setting of MCL \cite{Yue2023ModelContrastiveLF}, where we use the trigger prior from target synthesis literature \cite{Tao2022BetterTI}. The main idea is to use the trigger prior to get the location of poisoned prototypes ${P'}$ and apply our sanitization loss on this synthetically computed direction $V^s$. When using the large model mapping pipeline alongside ST-PGBD, mapping is done for both $\{P\}$ and $\{P'\}$. 
We show dataset-wise results in \Cref{tab:abl_variations}. We observe similar trends to the mean results where $V^s$ shows marginally better ASR reduction, but falls behind on overall DEM when compared to base PGBD.
\begin{figure*}[h]
    \centering
    \includegraphics[height=80mm]{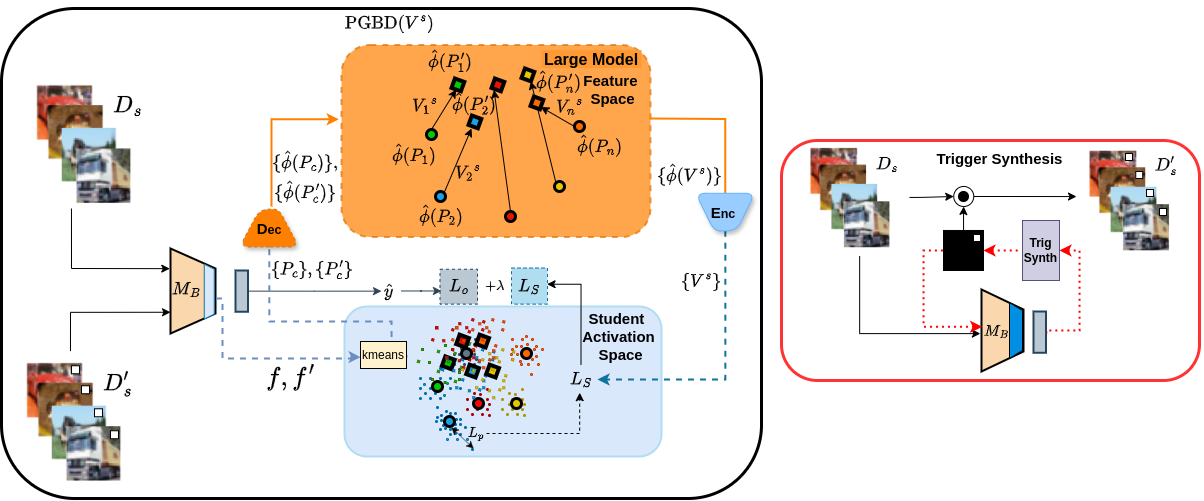}
    \caption{{\bf [Left]} Block diagram for ST-PGBD (Note that bold squares correspond to poisoned prototypes $P'_c$). {\bf [Right]} Trigger synthesis process given target label t \textbf{(right)}.}
    \label{fig:synth_block}
\end{figure*}
\textbf{Note:} We provide a video to explain our method and its variations in a step-by-step manner as part of the supplementary.
\section{Implementation Details}
\label{sec:supple_implementation}
In this section, we provide finer implementation details pertaining to the architecture and various experiments in the main paper.

\Paragraph{Trigger Synthesis:} 
We use the same configuration as MCL \cite{Yue2023ModelContrastiveLF} to synthesize the trigger using the optimization based method by \citet{Tao2022BetterTI}. 
Synthesis is done once for a given model, dataset, and attack combination and is reused for all the future runs with that combination. 

\Paragraph{Mapping Module:}
The mapping module is fully convolutional with the encoder and decoder parts having a single convolutional layer each.
We train the mapping module for 5 epochs on a small subset of the available clean data \(D_s\) ($\approx$ 150-300 data points).
We use a constant learning rate of 0.01 for the training and the standard reconstruction loss using mean square error.
We train the encoder to map from the teacher feature space to the student feature space and the decoder to map from the student feature space to the teacher \cite{caron2021emerging} feature space. Once trained, the mapping module parameters are updated only when the prototypes are updated and remain frozen at all other times.


\Paragraph{Defense phase:}
We use a learning rate of \(10^{-4}\) throughout our defense phase. We do not require any finetuning on \(D_s\) before our defense method, reducing the overall computation of the defense phase. We use an update gap of 5 epochs, \ie we update the prototypes, PAVs and mapping module every 5 epochs.  

\Paragraph{Testing phase:}
We implement the same pipeline as MCL for our evaluation but use the trigger implementations from BackdoorBench \cite{wu2022backdoorbench} when generating the poisoned test data. This is done to maintain consistency with the trigger in the train time attack. 

\subsection{Experiment Details}
\label{subsec:exp_details}

\Paragraph{Attack Setups:}
For the attacks, we use the code implementation by \citet{wu2022backdoorbench} to generate the backdoor models. 
An important point to note regarding the implementation of Trojan \cite{10.1145/3394486.3403064} attack is that we use a much larger trigger mask in our implementation as compared to previous works, which implies our implementation helps generate a stronger attack to defend against.
We do not show results on the Signal \cite{Barni2019ANB_signal} attack for CIFAR100, TinyImagenet, and GTSRB datasets since it is a clean label attack. Clean label attacks only poison the samples of the target class, whereas large datasets do not have enough samples per class to satisfy a poisoning rate of even 1\%.

\Paragraph{Defense Setups:}
For defenses, we use official implementations for MCL \cite{Yue2023ModelContrastiveLF} and NAD \cite{Li2021NeuralAD} and use the implementation of BackdoorBench \cite{wu2022backdoorbench} for the rest. We run each defense for 50 epochs in their respective default configurations and report the results on the best-performing model from all 50 epochs in terms of the overall DEM($\Gamma$) score.   

\Paragraph{Metrics:}
We define again the generic metrics CA (Clean Accuracy) and ASR (Attack success rate). Clean Accuracy is defined as the accuracy of the model on a clean test set. Attack Success Rate is defined as the accuracy of the model on a poisoned test set, where the labels are also swapped to the target class in the case of attacks requiring label perturbation. We already define our Defense Efficacy Measure (DEM) in the main paper (\Cref{eqn:DEM_linear}).

\Paragraph{Hyperparameters:}
Throughout the paper, we follow one base configuration irrespective of attack or dataset, which we detail below:\\
\emph{Base configuration:} For the base configuration we update CAVs every 5 epochs, with $\alpha=0.75$. We use no dropout and set $\delta = 0.75$. PGBD is run for 35 epochs, with a $D_s$ of 5\% size of the original training dataset. 

We provide changes to the hyperparameter configurations (if done) for all the results shown in the paper:
\begin{itemize}
   \item Table 2 (sunglasses): $\alpha=0.5$, $\lambda=25$
   \item Table 2 (mask): $\alpha=0.5$, $\delta=25$
   \item Table 2 (tattoo): $\alpha=0.25$, $\lambda=10$ 
   \item Table 4 (NT-PGBD): cycling interval = 3, total number of epochs = 30
\end{itemize}

\begin{table}[t]
\centering
\begin{tabular}{l||c|c|c}
\toprule[1.5pt]
\text{Attack} & \text{Poisoning} & \text{\#Filters} & \text{Avg Images} \\
 & \text{Rate (\%)} & \text{Used} & \text{per Class} \\
\midrule[0.5pt]
sunglasses  & 10 & NIL & 53 \\
tattoo & 5 & 1 & 41 \\
mask  & 10 & 4 & 41 \\
\bottomrule[1.5pt]
\end{tabular}
\caption{Hyper-parameters and Experiment Settings for different Semantic Attack Types. Sunglasses based dataset consists of only real-world images.}
\label{tab:combined_settings}
\end{table}

\Paragraph{Validation Setup:}
For the validation experiments in Sec3 of the main paper, we use 20\% of the training data as our sample. We get the backdoored models by training on the attack implementations of BackdoorBench \cite{wu2022backdoorbench} in our setting for generating ground truth poisoned images for our validation and ablation experiments. 

\section{Semantic Attack Dataset: Details}
\label{sec:supple_datasets}
we use CIFAR10 \cite{krizhevsky2009learning} (10 classes, 50000 train images, and 10000 test images), CIFAR100 \cite{krizhevsky2009learning} (100 classes, 50000 train, and 10000 test images), GTSRB \cite{STALLKAMP2012323} (43 classes, 39209 train, and 12630 test images) with the corresponding mentioned train/test splits.
In addition to these datasets, we create our own semantic face occlusion attack dataset from the publicly available ROF \cite{erakiotan2021recognizing} dataset. We provide dataset creation procedure in the following section.
\subsection{Semantic Face Occlusion Dataset}

The publicly available ROF dataset does not contain enough data for mask occlusion to properly test the effect of such an attack. To overcome the challenge of insufficient data, we created our own dataset by synthetically adding the occlusion onto the image. The occlusion based trigger was added via Snapchat filters which are very robust at face detection, even with low quality images. Onto each image, a filter was added which adds the corresponding object onto the face, aligned with the pose of the face. Filter was carefully added onto each image after centering the face properly and adjusting it slightly to look natural. We also add a tattoo based attack which help us understand how the size of the semantic occlusion also affects the performance of our method. The average number of images per class are lower for synthetic attacks is due to the extremely pixelated images present in ROF dataset of the celebrity face images which snapchat could not recognise as face. While defending we use the standard hyperparameters for MCL, except for tattoo, where we found a temperature of 0.3 to give best performance. We detail our hyperparameters in \Cref{subsec:exp_details}.
We give trigger-wise experiment details in \Cref{tab:combined_settings}. Additionally, we will provide the datasets created as part of the supplementary material.


\begin{table*}[h!]
\label{tab:abl_all2all}
\centering
\begin{tabular}{l||cc||ccc|ccc}
    \toprule[1.5pt]
    Method \faCaretRight & \multicolumn{2}{c||}{Baseline}   & \multicolumn{3}{c|}{PGBD} & \multicolumn{3}{c}{NT-PGBD} \\
    Target \faCaretDown & CA $\uparrow$ & ASR $\downarrow$ & CA $\uparrow$ & ASR $\downarrow$ & $\Gamma \uparrow$ & CA $\uparrow$ & ASR $\downarrow$ & $\Gamma \uparrow$ \\
\hline 
    Badnet & 89.58 & 85.06 & 89.93 & 1.75 & 0.99 & 91.17 & 1.04 & 0.99 \\
    Trojan & 93.43 & 81.51 & 89.75 & 5.26 & 0.95 & 87.22 & 10.48 & 0.90 \\
    Blended & 92.93 & 66.32 & 88.28 & 3.56 & 0.95 & 86.59 & 5.3 & 0.93 \\
\hline 
    \end{tabular}
\caption{Results for defense of the all-to-all attack configuration with PGBD and NT-PGBD on CIFAR10 dataset. NT-PGBD shows on par-performance to PGBD without knowledge of target class mappings.}
\end{table*}


\section{Additional Ablation Results}
\label{sec:ablations_supple}
\subsection{Attack time ablations}
We provide results and additional inferences for our attack time ablation analysis in this section.
\Paragraph{All-to-All attack:}
While base PGBD cannot infer an all-to-all attack on its own, it is possible to infer an all-to-all case through the trigger synthesis module used for obtaining the target class. Therefore, we propose two ways of defending against all-to-all attack scenario: 1) Assuming knowledge of the all-to-all configuration, update PAVs \(V^p\) to use the adjacent class as target class, or 2) Use the NT-PGBD approach that does need the target. We show results for both in \Cref{tab:abl_all2all}. Surprisingly, NT-PGBD performs better than PGBD which reinforces our claim of using it as a post-hoc brute force model recovery technique. We believe the  reason behind the success of NT-PGBD is that the trigger feature is learned as a part of every class in an all-to-all attack. Since NT-PGBD defends in a round robin manner, the trigger feature is effectively scrubbed from the model.

\Paragraph{Target label ablation:}
While our method considers the target class's position in the activation space, we show in \Cref{tab:abl_target_label} that our class-specific PAV choice enables us to work for any target class. We ablate the target class label and repeat the defense experiment for the CIFAR10 dataset and Badnet attack to validate this.
\Cref{tab:abl_target_label} shows that variation in the target class label (\ie the target class) has little effect on the overall performance with the standard deviation of DEM values = 0.015 only overall.  
\begin{table}[b]
\centering
    \begin{tabular}{c||cc||ccc}
    \toprule[1.5pt]
    Method \faCaretRight & \multicolumn{2}{c||}{Baseline}   & \multicolumn{3}{c}{PGBD} \\
    Target \faCaretDown & CA$\uparrow$ & ASR$\downarrow$ & CA $\uparrow$ & ASR$\downarrow$ & $\Gamma\uparrow$ \\
    \midrule[1.5pt]
    0 & 92.34 & 88.93 & 90.66 & 0.82 & 0.99 \\
    3 & 93.20 & 76.34 & 88.02 & 0.33 & 0.97 \\
    5 & 93.40 & 90.11 & 88.44 & 1.86 & 0.96 \\
    9 & 92.58 & 91.01 & 92.06 & 0.32 & 0.99 \\
    \bottomrule[1.5pt]
    \end{tabular}
\caption{Results on varying the target label for the Badnet attack with CIFAR10 dataset. PGBD is robust to target label changes.}
\label{tab:abl_target_label}
\end{table}

\begin{figure}[h]
    \centering
    \includegraphics[width=\linewidth,height=40mm]{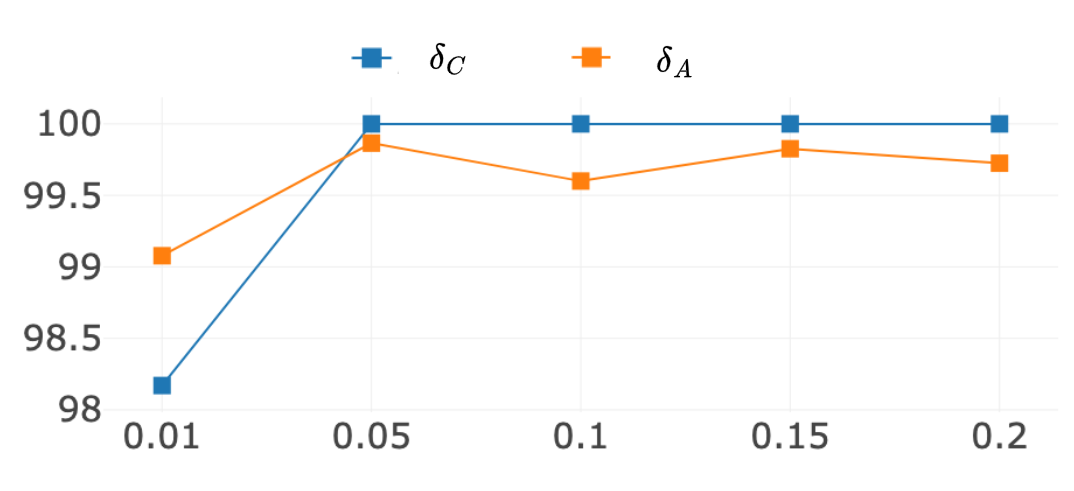}
    \caption{\(\delta_C\) and \(\delta_A\) graphs for our method with varying values of poisoning rate on the Badnet attack. For a successful defense, both $\delta_A$ and $\delta_C$ should be close to 1 (see \Cref{eqn:DEM_linear}).  PGBD is robust to even large poisoning rates of 20\%. }
    \label{fig:poison_rate_effect}
\end{figure}
\Paragraph{Poisoning Rate:} 
Poisoning rate is the amount of training dataset that the attacker modifies. 
A higher poisoning rate implies a more potent attack due to more exposure to the trigger feature at training time. Our method is successful at even high levels of poisoning of 20\% of the training data. We observe an increase in CA levels with near-constant levels of ASR reduction as we use higher poisoning rates.

\begin{table}[b]

    \begin{tabular}{C{20mm}||cc|ccc}
    \toprule[1.5pt]
    \multicolumn{1}{r||}{Method \faCaretRight} & \multicolumn{2}{c|}{Baseline}   & \multicolumn{3}{c}{PGBD} \\
    Target \faCaretDown & CA$\uparrow$ & ASR$\downarrow$ & CA$\uparrow$ & ASR$\downarrow$ & $\Gamma\uparrow$\\
\hline 
    PreactResNet18 & 92.35 & 88.93 & 90.66 & 0.82 & 0.986 \\
    PreactResNet34 & 93.22 & 87.69 & 92.46 & 0.08 & 0.995 \\
    PreactResNet50 & 93.07 & 92.32 & 93.76 & 0.3 & 0.998 \\
\hline 
    \end{tabular}
    \caption{Uniform results on Badnet and CIFAR10 with varying model sizes using PreactResNet architecture for PGBD.}
\label{tab:abl_model_sz}
\end{table}

\Paragraph{Effect of model size}
To understand if there is any effect of an increase in model size on the performance of our defense, we run our method on two larger variations of pre-act ResNet, namely pre-act ResNet34 and pre-act ResNet50. We show the results of our method on the Badnet attack on CIFAR10 for three model size variations in \Cref{tab:abl_model_sz}. We observe uniform performance overall model sizes implying that our defense robustly scales for larger models.

\Paragraph{Ablation on model architectures:} We show results for all attacks on CIFAR10 with the VGG architecture(VGG19\_bn) in \Cref{tab:reb_vgg19}. We observe that performance is not on par with that observed in ResNet architecture, implying that activation space dependent methods are affected by model architecture. Nevertheless, PGBD achieves state of the art performance on the VGG architecture as well. 
\begin{table*}
\centering
\begin{tabular}{l||cc||ccc||ccc|ccc}
\toprule[1.2pt]

Method  & \multicolumn{2}{c||}{Baseline} & \multicolumn{3}{c||}{MCL} & \multicolumn{3}{c||}{ST-PGBD} & \multicolumn{3}{c}{PGBD}  \\
\midrule
Attack  & CA & ASR & CA & ASR & $\Gamma$ & CA & ASR & $\Gamma$ & CA & ASR & $\Gamma$    \\
\specialrule{0.75pt}{1\jot}{1\jot}
Badnet &  {90.68} &  {87.00} & 82.94 & 4.02 & 0.93 & 90.98 & 6.99 &  \underline{0.96} & 89.69 & 1.51 & \textbf{0.99} \\
Trojan & 91.16 & 100.00 &  {88.32} &  {6.48} &  \underline{0.95} & 71.40 & 0.51 & 0.89 &  {89.60} & 4.80 & \textbf{0.97} \\
Blended & 90.89 &  {85.44} &  {74.48} &  {71.87} & 0.49 & 77.70 &  {28.10} &  \underline{0.76} &  {77.19} &  {9.49} & \textbf{0.87} \\
Signal & 91.94 & 99.93 & 79.47 &  {65.13} & 0.61 &  {88.82} &  {8.68} &  \underline{0.94} &  {88.39} & 5.90 & \textbf{0.95} \\
Wanet & 85.22 & 91.08 & 80.12 & 13.21 & 0.90 & 88.67 & 1.98 &  \underline{0.99} & 90.85 & 0.45 & \textbf{1.00}\\

\bottomrule[1.2pt]
\end{tabular}
\caption{Results with VGG19 model for all attacks on CIFAR10.}
\label{tab:reb_vgg19}
\end{table*}

\subsection{Defense time ablations} \label{subsec:def_abl}
\begin{figure}
    \centering
    \includegraphics[width=\linewidth]{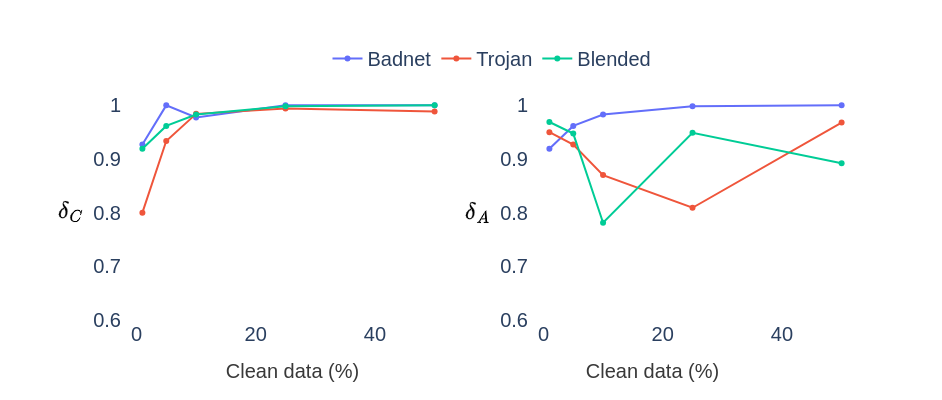}
    \caption{{\bf [Left]} $\delta_C$ and {\bf [Right]} $\delta_A$ graphs for our method with varying percentages of clean data on three attacks. We are robust to available clean data size across attacks, unlike MCL\cite{Yue2023ModelContrastiveLF}.}
    \label{fig:clean_data_size}
\end{figure}

\Paragraph{Clean Data Availability:}
The effect on the defense performance on varying the size of available clean data is shown in \Cref{fig:clean_data_size}. 
For PGBD, we observe no adverse effect on CA retention and nearly uniform performance in terms of ASR reduction. This highlights the robustness of our system compared to MCL, which faces CA drop for $>=$20\% availability because of their dependence on negative pairs in contrastive loss.

\subsection{Design Analysis}\label{abl:DesAnal}
\Paragraph{Prototype Computation ablation}
As a preliminary analysis of alternative prototype computation methods, we used k-medoids instead of k-means as the clustering technique for prototype computation. We observed no significant performance difference as shown in \cref{tab:abl_kmedoids}. This indicates that our modified prototype computation process is robust due to the final averaging step, which condenses the prototype down to a single point in the activation space, where variations in individual cluster centers of a class are largely neutralized. 
\begin{table}
\centering
\small
\begin{tabular}{l||cc||ccc|ccc}
\toprule[1.2pt]
 Method & \multicolumn{2}{c||}{\textbf{Baseline}} & \multicolumn{3}{c|}{\textbf{PGBD} (k-means)}      & \multicolumn{3}{c}{\textbf{PGBD} (k-medoids)} \\
 \midrule[1pt]
Attack & CA  & ASR  & CA  & ASR  & $\Gamma$ & CA  & ASR & $\Gamma$  \\
\midrule[1.5pt]
Badnet  & 92.34 & 88.93  & 90.66 & 0.82 & 0.99 & 89.43 & 1.51 & 0.98 \\
Trojan  & 93.00 & 100.00 & 83.60 & 6.76 & 0.92 & 85.95 & 8.2 & 0.92 \\
Blended & 93.06 & 92.94  & 86.11 & 4.87 & 0.94 & 85.16 & 7.63 & 0.92 \\
Sig     & 92.90 & 89.04  & 87.18 & 0.31 & 0.97 & 87.17 & 2.77 & 0.95 \\
Wanet   & 89.98 & 97.60  & 88.54 & 2.36 & 0.98 & 88.22 & 2.44 & 0.98 \\
\bottomrule[1.2pt]
\end{tabular}
\caption{Results on using k-medoids instead of k-means for class-wise prototype computation on CIFAR10 dataset with the default value of k=3 for both cases. No significant change in performance is observed.}
\label{tab:abl_kmedoids}
\end{table}
Additionally, we varied $k$ across attacks for prototype computation in \cref{tab:abl_kmeans_k}. Our choice of $k$ = 3 proves to be the right choice
overall; a slight degradation in performance was observed in the case of Trojan and Wanet attacks when using large $k$ (=20).
\begin{table*}
\label{tab:abl_kmeans_k}
\small
\centering
\begin{tabular}{l||cc||ccc|ccc|ccc|ccc}
\toprule[1.2pt]

Method \faCaretRight & \multicolumn{2}{c||}{Baseline} & \multicolumn{3}{c|}{PGBD ($k$=3)} & \multicolumn{3}{c|}{PGBD ($k$=5)} & \multicolumn{3}{c|}{PGBD ($k$=10)} & \multicolumn{3}{c}{PGBD ($k$=20)} \\
Attack \faCaretDown & CA & ASR & CA & ASR & $\Gamma$ & CA & ASR & $\Gamma$ & CA & ASR & $\Gamma$ & CA & ASR & $\Gamma$  \\
\hline 
Badnet & 92.34 & 88.93 & 90.66 & 0.82 & \textbf{0.99} & 88.59 & 1.36 & \underline{0.97} & 88.27 & 1.15 & \underline{0.97} & 88.19 & 1.71 & \underline{0.97} \\
Trojan & 93.00 & 100.00 & 83.60 & 6.76 & \textbf{0.92} & 83.60 & 12.14 & \underline{0.89} & 88.83 & 45.54 & 0.75 & 81.08 & 35.69 & 0.76 \\
Blended & 93.06 & 92.94 & 86.11 & 4.87 & \textbf{0.94} & 87.19 & 12.13 & 0.90 & 81.00 & 5.02 & \underline{0.91} & 86.10 & 4.02 & \textbf{0.94} \\
Sig & 92.90 & 89.04 & 87.18 & 0.31 & \textbf{0.97} & 89.29 & 1.17 & \textbf{0.97} & 89.85 & 3.36 & \underline{0.96} & 85.58 & 0.91 & \underline{0.96} \\
Wanet & 89.98 & 97.60 & 88.54 & 2.36 & \underline{0.98} & 90.48 & 2.41 & \textbf{0.99} & 87.01 & 3.10 & 0.97 & 85.63 & 2.39 & 0.96\\
\bottomrule[1.2pt]
\end{tabular}
\caption{Results on varying values of $k$ on CIFAR10 dataset}
\end{table*}

\Paragraph{Ground truth trigger ablation}
We perform ablation for \(V^s\) where we use the ground truth trigger (giving us \(V^{s^{*}}\)) instead of the synthesized trigger (Table S2). It should be noted that we observe much better performance against signal attack when using ground truth trigger in comparison to the synthesized trigger. 
This is due to the shortcomings of the trigger synthesis optimization \cite{Tao2022BetterTI}, which is unable to reconstruct signal-based triggers properly.
This validates our previous reasoning of error/failure propagation through the defense pipeline for the overly trigger synthesis-dependent works like \cite{Yue2023ModelContrastiveLF}.
\begin{table*}[h]

\centering
    \begin{tabular}{@{}L{20mm}||C{10mm}C{10mm}||C{10mm}C{10mm}C{10mm}|C{10mm}C{10mm}C{10mm}@{}}
    \toprule[1.5pt]
    METHOD \faCaretRight & \multicolumn{2}{c||}{Baseline}   & \multicolumn{3}{c|}{ST-PGBD} & \multicolumn{3}{c}{ST-PGBD with $V^{s^*}$} \\
    \midrule[0.5pt]
    TARGET \faCaretDown & CA $\uparrow$ & ASR $\downarrow$ & CA $\uparrow$ & ASR $\downarrow$ & $\Gamma \uparrow$ & CA $\uparrow$ & ASR $\downarrow$ & $\Gamma \uparrow$ \\
    \midrule[1.5pt]
    Badnet & 92.35 & 88.93 & 91.22 & 1.70 & 0.984 & 90.43 & 0.54 & 0.987 \\
    Blended & 93.06 & 92.94 & 87.51 & 2.58 & 0.956 & 87.91 & 3.14 & 0.955 \\
    Signal & 92.90 & 89.04 & 89.94 & \textcolor{cyan}{7.51} & 0.942 & 89.13 & \textcolor{cyan}{1.12} & 0.973 \\
    Wanet & 89.98 & 97.6 & 83.61 & 0.61 & 0.962 & 86.04 & 0.85 & 0.974 \\
    Trojan & 93.00 & 100.0 & 76.34 & 2.68 & 0.897 & 78.29 & 2.51 & 0.908 \\
    \bottomrule[1.5pt]
    
    \end{tabular}
\caption{Results on CIFAR10 when using ground truth trigger ($V^{s^{*}}$)}
\label{tab:abl_trig}
\end{table*}

\Paragraph{Effect of $\lambda$:}
\begin{figure}
    \centering
    \includegraphics[width=\linewidth]{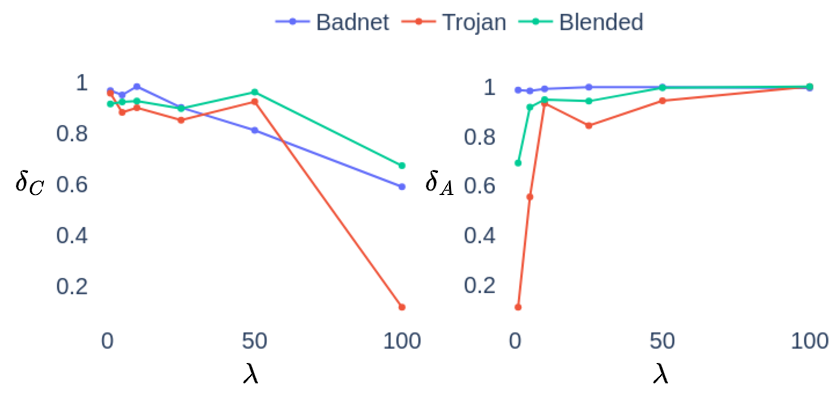}
    \caption{Effect of varying $\lambda$ in Eqn4 (main paper). Ideally, $\delta_C$ and $\delta_A$ must be 1 (see Eqn5 in main paper). We observe that increasing $\lambda$ decreases CA retention while increasing ASR reduction. Note, $\lambda = 0$ corresponds to naive finetuning.}
    \label{fig:abl_lambda}
\end{figure}
We observe the changes in $\delta_A$ and $\delta_C$ on changing the value of $\lambda$ in \Cref{fig:abl_lambda}. Simple finetuning is reperesnted by $\lambda$ = 0, and increasing $\lambda$ implies greater weightage given to $L_s$. We observe that $\lambda \in [1,50]$ give desirable results. Ideally, $\lambda$ should be chosen such that the magnitudes of $L_o$ and $L_s$ are comparable, since we would like to give equal weightage overall to both criteria. However, for harder attack to defend against, increasing $\lambda$ gives better results, like Trojan at $\lambda = 50$.

\begin{figure}[b]
    \centering
    \includegraphics[width=\linewidth]{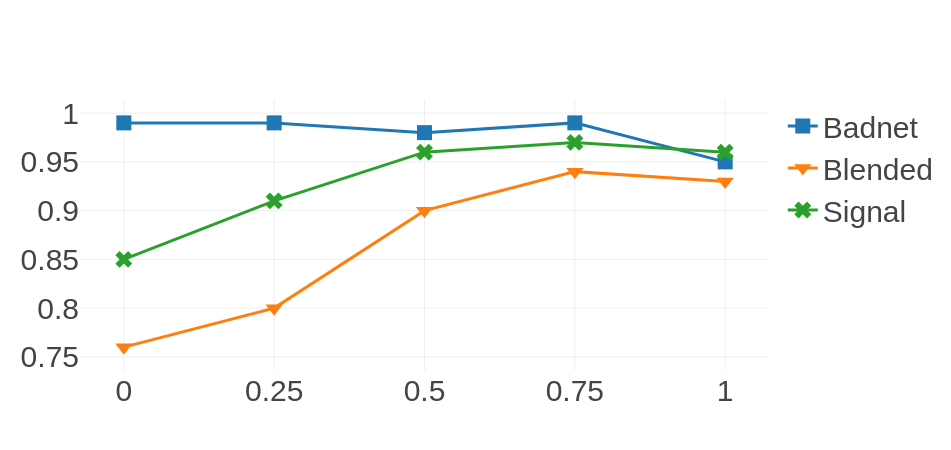} \\
    \caption{Line plots of DEM values for different attacks on CIFAR10 with varying $\alpha$ values.\vspace*{\baselineskip}}
    \label{fig:abl_alpha}
\end{figure}

\Paragraph{Alpha Ablation}
We use $\alpha$ to set the extent of update to be done when recomputing prototypes. We observe that smaller $\alpha$ values lead to better retention of CA after updates but at the cost of loss of overall ASR drop. This shortcomings of constant $\alpha$ can be seen in the results of Blended in \Cref{fig:abl_alpha}. We also show empirical reasoning for our choice of \(\alpha = 0.75\) through the figure.   
\begin{table*}
\centering
    \begin{tabular}{@{}L{18mm}||C{9mm}C{9mm}||C{9mm}C{9mm}C{9mm}|C{9mm}C{9mm}C{9mm}||C{9mm}C{9mm}C{9mm}|C{9mm}C{9mm}C{9mm}@{}}
    \toprule[1.5pt]
    METHOD \faCaretRight & \multicolumn{2}{c||}{Baseline} & \multicolumn{3}{c|}{PGBD, \(\alpha = 0.75\)} & \multicolumn{3}{c||}{PGBD, \(\alpha\) schd.} & \multicolumn{3}{c|}{ST-PGBD, $\alpha = 0.75$} & \multicolumn{3}{c}{ST-PGBD, \(\alpha\) schd.} \\ 
    \midrule[0.5pt]
     ATTACK \faCaretDown & CA & ASR & CA & ASR & $\Gamma$ & CA & ASR & $\Gamma$ & CA & ASR & $\Gamma$ & CA & ASR & $\Gamma$ \\ 
     \midrule[1.5pt]
    Badnet & {92.34} & 88.93 & {90.66} & {0.82} & \underline{0.99} & {92.80} & {0.18} & \textbf{1.00} & {91.22} & {1.70} & 0.98 & {90.87} & {0.70} & \underline{0.99} \\
    Trojan & {93.00} & 100.0 & {83.60} & {6.76} & \underline{0.92} & {87.22} & {18.30} & 0.88 & {76.34} & {2.68} & 0.90 & {89.92} & {6.78} & \textbf{0.95} \\
    Blended & {93.06} & 92.94 & {86.11} & {4.87} & \underline{0.94} & {89.84} & {12.43} & 0.92 & {87.51} & {2.58} & \textbf{0.96} & {90.56} & {9.67} & 0.93 \\
    Sig & {92.90} & 89.04 & {87.18} & {0.31} & \textbf{0.97} & {88.08} & {0.91} & \textbf{0.97} & {89.94} & {7.51} & 0.94 & {87.38} & {3.60} & \underline{0.95} \\
    Wanet & {89.98} & 97.60 & {88.54} & {2.36} & \textbf{0.98} & {85.91} & {5.54} & 0.95 & {83.61} & {0.61} & \underline{0.96} & {89.80} & {3.32} & \textbf{0.98} \\ 
    \bottomrule[1.5pt]
    \end{tabular}
    \caption{Results with $\alpha$ scheduler for CIFAR10}
\label{tab:sched_CIFAR10}
\end{table*}
\;\newline
\begin{table*}
\centering
    \begin{tabular}{@{}L{18mm}||C{9mm}C{9mm}||C{9mm}C{9mm}C{9mm}|C{9mm}C{9mm}C{9mm}||C{9mm}C{9mm}C{9mm}|C{9mm}C{9mm}C{9mm}@{}}
    \toprule[1.5pt]
    METHOD \faCaretRight & \multicolumn{2}{c||}{Baseline} & \multicolumn{3}{c|}{PGBD, \(\alpha = 0.75\)} & \multicolumn{3}{c||}{PGBD, \(\alpha\) schd.} & \multicolumn{3}{c|}{ST-PGBD, \(\alpha = 0.75\)} & \multicolumn{3}{c}{ST-PGBD, \(\alpha\) schd.} \\ \midrule[0.5pt]
     ATTACK \faCaretDown & CA & ASR & CA & ASR & $\Gamma$ & CA & ASR & $\Gamma$ & CA & ASR & $\Gamma$ & CA & ASR & $\Gamma$ \\
     \midrule[1.5pt]
    Badnet & {67.32} & 86.98 & {63.29} & {0.01} & \textbf{0.97} & {62.63} & {0.00} & \textbf{0.97} & {62.80} & {0.03} & \textbf{0.97} & {62.62} & {0.00} & \textbf{0.97} \\ 
    Trojan & {70.02} & 100.0 & {60.01} & {0.00} & \textbf{0.93} & {60.01} & {0.00} & \textbf{0.93} & {54.53} & {0.04} & 0.89 & {59.34} & {0.01} & \underline{0.92} \\ 
    Blended & {69.01} & 99.48 & {58.49} & {0.06} & \textbf{0.92} & {57.38} & {0.16} & \underline{0.91} & {42.89} & {2.03} & 0.80 & {58.30} & {44.16} & 0.70 \\ 
    Wanet & {63.84} & 91.47 & {59.64} & {0.01} & \textbf{0.97} & {60.33} & {0.48} & \textbf{0.97} & {57.04} & {0.90} & 0.94 & {56.33} & {1.03} & \underline{0.94} \\ 
    \bottomrule[1.5pt]
    \end{tabular}
    \caption{Results with $\alpha$ scheduler for CIFAR100}
\label{tab:sched_CIFAR100}
\end{table*}


\begin{table*}
\centering
    \begin{tabular}{@{}L{18mm}||C{9mm}C{9mm}||C{9mm}C{9mm}C{9mm}|C{9mm}C{9mm}C{9mm}||C{9mm}C{9mm}C{9mm}|C{9mm}C{9mm}C{9mm}@{}}
    \toprule[1.5pt]
    METHOD \faCaretRight & \multicolumn{2}{c||}{Baseline} & \multicolumn{3}{c|}{PGBD, $\alpha = 0.75$} & \multicolumn{3}{c||}{PGBD, $\alpha$ schd.} & \multicolumn{3}{c|}{ST-PGBD, $\alpha = 0.75$} & \multicolumn{3}{c}{ST-PGBD, $\alpha$ schd.} \\ \midrule[0.5pt]
     ATTACK \faCaretDown & CA & ASR & CA & ASR & $\Gamma$ & CA & ASR & $\Gamma$ & CA & ASR & $\Gamma$ & CA & ASR & $\Gamma$ \\
     \midrule[1.5pt]
    Badnet & {96.61} & 83.86 & {97.26} & {0} & \textbf{1.00} & {97.45} & {0.04} & \textbf{1.00} & {96.79} & {0.00} & \textbf{1.00} & {97.74} & {0.02} & \textbf{1.00} \\ 
    Trojan & {98.17} & 100.0 & {96.5} & {0.11} & \textbf{0.99} & {95.80} & {1.25} & \underline{0.98} & {97.18} & {0.06} & \textbf{0.99} & {96.27} & {0.02} & \textbf{0.99} \\ 
    Blended & {98.66} & 96.33 & {86.2} & {0.72} & 0.93 & {95.24} & {0.00} & \textbf{0.98} & {89.94} & {3.67} & 0.94 & {96.73} & {3.08} & \underline{0.97} \\ 
    Wanet & {98.04} & 82.14 & {97.14} & {0.33} & \underline{0.99} & {97.78} & {0.57} & \textbf{1.00} & {97.35} & {0.08} & \textbf{1.00} & {98.08} & {0.01} & \textbf{1.00} \\ 
    \bottomrule[1.5pt]
    \end{tabular}
    \caption{Results with $\alpha$ scheduler for GTSRB}
\label{tab:sched_GTSRB}
\end{table*}
\begin{table*}
    \begin{tabular}{@{}L{18mm}||C{9mm}C{9mm}||C{9mm}C{9mm}C{9mm}|C{9mm}C{9mm}C{9mm}||C{9mm}C{9mm}C{9mm}|C{9mm}C{9mm}C{9mm}@{}}
    \toprule[1.5pt]
    METHOD \faCaretRight & \multicolumn{2}{c||}{Baseline} & \multicolumn{3}{c|}{PGBD, $\alpha = 0.75$} & \multicolumn{3}{c||}{PGBD, $\alpha$ schd.} & \multicolumn{3}{c|}{ST-PGBD, $\alpha = 0.75$} & \multicolumn{3}{c}{ST-PGBD, $\alpha$ schd.} \\ \midrule[0.5pt]
     ATTACK \faCaretDown & CA & ASR & CA & ASR & $\Gamma$ & CA & ASR & $\Gamma$ & CA & ASR & $\Gamma$ & CA & ASR & $\Gamma$ \\
     \midrule[1.5pt]
    Badnet & {85.42} & 86.59 & {83.74} & {0.28} & \textbf{0.99} & {84.29} & {0.07} & \textbf{0.99} & {83.60} & {0.58} & 0.98 & {83.74} & {0.24} & \textbf{0.99} \\ 
    Trojan & {87.06} & 100.0 & {80.04} & {2.29} & \underline{0.95} & {81.01} & {6.52} & 0.93 & {76.02} & {0.93} & 0.93 & {81.84} & {2.27} & \textbf{0.96} \\ 
    Blended & {86.91} & 96.25 & {76.93} & {1.88} & \underline{0.93} & {80.82} & {4.20} & \textbf{0.94} & {73.45} & {2.76} & 0.90 & {81.86} & {18.97} & 0.87 \\ 
    Sig & {92.90} & 89.04 & {87.18} & {0.31} & \textbf{0.97} & {88.08} & {0.91} & \textbf{0.97} & {89.94} & {7.51} & 0.94 & {87.38} & {3.60} & \underline{0.95} \\ 
    Wanet & {83.95} & 90.40 & {81.77} & {0.90} & \textbf{0.98} & {81.34} & {2.20} & \underline{0.97} & {79.58} & {0.51} & \underline{0.97} & {81.40} & {1.45} & \textbf{0.98} \\ 
    \bottomrule[1.5pt]
    \end{tabular}
    \caption{Mean results with $\alpha$ scheduler over all datasets}
\label{tab:sched_mean}
\end{table*}
\Paragraph{Alpha scheduler}
As observed previously, a single \(\alpha\) value throughout the sanitization does not always give the desired results. Therefore, we implemented a scheduler for \(\alpha\) through empirical observations of epoch-level changes in CA and ASR. For the scheduler, we set \(\alpha = 1\) for the first update, \(\alpha = 0.75\) for the second update, and \(\alpha = 0.5\) for the rest of the updates. 

We observe on-par or better results when using a scheduler with \(V^s\) and minor improvements in performance against signal attack in the case of \(V^p\) (\Cref{tab:sched_mean}). We also observe improvement in performance against Trojan when using the scheduler as seen from \Cref{fig:dem_all} (e).

\Paragraph{Mapping pipeline ablation} 
In the pipeline shown in \Cref{fig:synth_block} and in \Cref{fig:blockDiag}, we see that we perform large model mapping of only the classwise prototypes and not the complete dataset(\(D_s , D'_s\)). Our reasoning for doing this was to save computing time. Given that we use a very small mapping module, we expected minimal change in final outcome. As an ablation, we look at the results in \Cref{tab:abl_pipeline} when performing the prototype calculation step also in the teacher space, using \(\phi(D_s)\). 
\begin{table*}
\centering
    \begin{tabular}{@{}L{20mm}||C{10mm}C{10mm}||C{10mm}C{10mm}C{10mm}|C{10mm}C{10mm}C{10mm}@{}}
    \toprule[1.5pt]
    Method \faCaretRight & \multicolumn{2}{c||}{Baseline}   & \multicolumn{3}{c|}{ST-PGBD} & \multicolumn{3}{c}{ST-PGBD with $V^{s^*}$} \\
    \midrule[0.5pt]
    Attack \faCaretDown & CA $\uparrow$ & ASR $\downarrow$ & CA $\uparrow$ & ASR $\downarrow$ & $\Gamma \uparrow$ & CA $\uparrow$ & ASR $\downarrow$ & $\Gamma \uparrow$ \\
    \midrule[1.5pt]
    Badnet & 92.35 & 88.93 & 90.66 & 0.82 & 0.99 & 92.46 & 0.05 & 1.00 \\
Blended & 93.06 & 92.94 & 83.6 & 6.76 & 0.91 & 89.17 & 46.12 & 0.73 \\
Signal & 92.9 & 89.04 & 86.11 & 4.87 & 0.94 & 88.8 & 80.26 & 0.53 \\
Wanet & 89.98 & 97.6 & 87.18 & 0.31 & 0.98 & 87.75 & 14.19 & 0.91 \\
Trojan & 93 & 100 & 88.54 & 2.36 & 0.96 & 88.6 & 71.49 & 0.62 \\
    \bottomrule[1.5pt]
    
    \end{tabular}
    \caption{Results on CIFAR10 with altered pipeline described in \Cref{abl:DesAnal} (Mapping pipeline ablation). We do not observe significant changes in results.}
\label{tab:abl_pipeline}
\end{table*}

\Paragraph{Ablation with DINOv2}
We use DinoV1 to maintain the same setting as \cite{Gupta2023ConceptDL}. While using different recently available foundational models is an interesting experiment, our expectation from the foundational teacher model was to use its rich feature space, which is common to all available pre-trained foundational models. We show results when using DinoV2 \cite{oquab2023dinov2} in \Cref{tab:abl_dinov2} to show the consistency of our method in this aspect.


\begin{table*}
\label{tab:abl_dinov2}
\small
\centering
\begin{tabular}{l||cc||ccc|ccc||ccc|ccc}
\toprule[1.2pt]

Method \faCaretRight & \multicolumn{2}{c||}{Baseline} & \multicolumn{3}{c|}{PGBD with DinoV1.} & \multicolumn{3}{c||}{PGBD with DinoV2.} & \multicolumn{3}{c|}{ST-PGBD with DinoV1} & \multicolumn{3}{c}{ST-PGBD with DinoV2.} \\
Attack \faCaretDown & CA & ASR & CA & ASR & $\Gamma$ & CA & ASR & $\Gamma$ & CA & ASR & $\Gamma$ & CA & ASR & $\Gamma$  \\
\hline 
Badnet & 85.42 & 86.59 & 83.74 & 0.28 & 0.99 & 82.43 & 0.04 & 0.98 & 83.60 & 0.58 & 0.98 & 82.92 & 0.45 & 0.98 \\
Trojan & 87.06 & 100.00 & 80.04 & 2.29 & 0.95 & 81.61 & 0.43 & 0.96 & 76.02 & 0.93 & 0.93 & 81.88 & 0.60 & 0.96 \\
Blended & 86.91 & 96.25 & 76.93 & 1.88 & 0.93 & 80.71 & 1.23 & 0.95 & 73.45 & 2.76 & 0.90 & 79.63 & 12.60 & 0.89 \\
Sig & 92.90 & 89.04 & 87.18 & 0.31 & 0.97 & 90.06 & 0.02 & 0.98 & 89.94 & 7.51 & 0.94 & 85.50 & 0.34 & 0.96 \\
Wanet & 83.95 & 90.40 & 81.77 & 0.90 & 0.98 & 80.64 & 2.32 & 0.96 & 79.33 & 0.53 & 0.97 & 80.15 & 2.49 & 0.96\\
\bottomrule[1.2pt]
\end{tabular}
\caption{Mean results with DinoV2 as teacher for large-model mapping compared to DinoV1}
\end{table*}

\Paragraph{Large model mapping ablation} 
Extending the findings shown in the main paper, we run our method without using the teacher model on the other two datasets as well. We observe that in datasets with many classes, mapping helps tremendously in safeguarding clean accuracy, as seen in \Cref{tab:distill_GTSRB} and \Cref{tab:distill_CIFAR100}
\begin{table*}
\footnotesize
\centering
    \begin{tabular}{@{}L{18mm}||C{9mm}C{9mm}||C{9mm}C{9mm}C{9mm}|C{9mm}C{9mm}C{9mm}||C{9mm}C{9mm}C{9mm}|C{9mm}C{9mm}C{9mm}@{}}
    \toprule[1.5pt]
    Method \faCaretRight & \multicolumn{2}{c||}{Baseline} & \multicolumn{3}{c|}{PGBD with mapping} & \multicolumn{3}{c||}{PGBD w/o mapping} & \multicolumn{3}{c|}{ST-PGBD with mapping} & \multicolumn{3}{c}{ST-PGBD w/o mapping} \\
    \midrule[0.5pt]
     Attack \faCaretDown & CA & ASR & CA & ASR & $\Gamma$ & CA & ASR & $\Gamma$ & CA & ASR & $\Gamma$ & CA & ASR & $\Gamma$ \\ 
     \midrule[1.5pt]
Badnet & 92.35 & 88.93 & 90.66 & 0.82 & 0.99 & 73.18 & 0.34 & 0.89 & 91.22 & 1.7 & 0.98 & 82.89 & 0.6 & 0.95 \\
Trojan & 93 & 100 & 83.6 & 6.76 & 0.92 & 73.34 & 6.7 & 0.86 & 76.34 & 2.68 & 0.90 & 83.56 & 0.1 & 0.95 \\
Blended & 93.06 & 92.94 & 86.11 & 4.87 & 0.94 & 81.57 & 0.08 & 0.94 & 87.51 & 2.58 & 0.96 & 75.29 & 0.02 & 0.90 \\
Signal & 92.9 & 89.04 & 87.18 & 0.31 & 0.97 & 71.64 & 0.478 & 0.88 & 89.94 & 7.51 & 0.94 & 82.14 & 0.02 & 0.94 \\
Wanet & 89.98 & 97.6 & 88.54 & 2.36 & 0.98 & 86.9 & 0.09 & 0.98 & 83.61 & 0.61 & 0.96 & 81.1 & 0.01 & 0.95 \\ \midrule[0.5pt]
MEAN & 92.26 & 93.7 & 87.22 & 3.02 & 0.96 & 77.33 & 1.54 & 0.91 & 85.72 & 3.02 & 0.95 & 81 & 0.15 & 0.94\\
    \bottomrule[1.5pt]
    \end{tabular}
    \caption{Results with and without large-model mapping for CIFAR10}
\label{tab:distill_CIFAR10}
\end{table*}
\;\newline
\begin{table*}

\centering
\footnotesize
    \begin{tabular}{@{}L{18mm}||C{9mm}C{9mm}||C{9mm}C{9mm}C{9mm}|C{9mm}C{9mm}C{9mm}||C{9mm}C{9mm}C{9mm}|C{9mm}C{9mm}C{9mm}@{}}
    \toprule[1.5pt]
    Method \faCaretRight & \multicolumn{2}{c||}{Baseline} & \multicolumn{3}{c|}{PGBD with mapping} & \multicolumn{3}{c||}{PGBD w/o mapping} & \multicolumn{3}{c|}{ST-PGBD with mapping} & \multicolumn{3}{c}{ST-PGBD w/o mapping} \\ \midrule[0.5pt]
     Attack \faCaretDown & CA & ASR & CA & ASR & $\Gamma$ & CA & ASR & $\Gamma$ & CA & ASR & $\Gamma$ & CA & ASR & $\Gamma$ \\
     \midrule[1.5pt]
Badnet & 67.32 & 86.98 & 63.29 & 0.01 & 0.97 & 55.25 & 0.22 & 0.91 & 62.80 & 0.03 & 0.97 & 53.12 & 0.03 & 0.89 \\
Trojan & 70.02 & 100.00 & 60.01 & 0.00 & 0.93 & 49.04 & 0 & 0.85 & 54.53 & 0.04 & 0.89 & 51.25 & 0.01 & 0.87 \\
Blended & 69.01 & 99.48 & 58.49 & 0.06 & 0.92 & 52.41 & 0.77 & 0.88 & 42.89 & 2.03 & 0.80 & 46.9 & 0 & 0.84 \\
Wanet & 63.84 & 91.47 & 59.64 & 0.01 & 0.97 & 55.19 & 0.49 & 0.93 & 57.04 & 0.90 & 0.94 & 57.91 & 0.03 & 0.95 \\ \midrule[0.5pt]
MEAN & 67.55 & 94.48 & 60.36 & 0.02 & 0.95 & 52.97 & 0.37 & 0.89 & 54.32 & 0.75 & 0.90 & 52.29 & 0.02 & 0.89 \\
    \bottomrule[1.5pt]
    \end{tabular}
    \caption{Results with and without large-model mapping for CIFAR100}
\label{tab:distill_CIFAR100}
\end{table*}
\;\newline\newline
\begin{table*}
\footnotesize
\centering
    \begin{tabular}{@{}L{18mm}||C{9mm}C{9mm}||C{9mm}C{9mm}C{9mm}|C{9mm}C{9mm}C{9mm}||C{9mm}C{9mm}C{9mm}|C{9mm}C{9mm}C{9mm}@{}}
    \toprule[1.5pt]
    Method \faCaretRight & \multicolumn{2}{c||}{Baseline} & \multicolumn{3}{c|}{PGBD with mapping} & \multicolumn{3}{c||}{PGBD w/o mapping} & \multicolumn{3}{c|}{ST-PGBD with mapping} & \multicolumn{3}{c}{ST-PGBD w/o mapping} \\ \midrule[0.5pt]
     Attack \faCaretDown & CA & ASR & CA & ASR & $\Gamma$ & CA & ASR & $\Gamma$ & CA & ASR & $\Gamma$ & CA & ASR & $\Gamma$ \\
     \midrule[1.5pt]
Badnet & 96.61 & 83.86 & 97.26 & 0.00 & 1.00 & 92.53 & 0 & 0.98 & 96.79 & 0.00 & 1.00 & 96.16 & 0 & 1.00 \\
Trojan & 98.17 & 100.00 & 96.50 & 0.11 & 0.99 & 88.83 & 1.3 & 0.95 & 97.18 & 0.06 & 0.99 & 95.21 & 0.85 & 0.98 \\
Blended & 98.66 & 96.33 & 86.20 & 0.72 & 0.93 & 87.0 & 0.45 & 0.94 & 89.94 & 3.67 & 0.94 & 85.55 & 1.29 & 0.93 \\
Wanet & 98.04 & 82.14 & 97.14 & 0.33 & 0.99 & 96.1 & 0 & 0.99 & 97.35 & 0.08 & 1.00 & 94.39 & 0 & 0.98 \\ \midrule[0.5pt]
MEAN & 97.87 & 90.58 & 94.28 & 0.29 & 0.98 & 91.12 & 0.44 &	0.96 & 95.32 & 0.95 & 0.98 & 92.83 & 0.54 & 0.97 \\
    \bottomrule[1.5pt]
    \end{tabular}
    \caption{Results with and without large-model mapping for GTSRB.}
\label{tab:distill_GTSRB}
\end{table*}

\section{Additional Result Visualizations}
\label{sec:supple_visualizations}
\subsection{t-SNE plots}
\begin{figure*}
    \centering
    \includegraphics[width=0.9\linewidth]{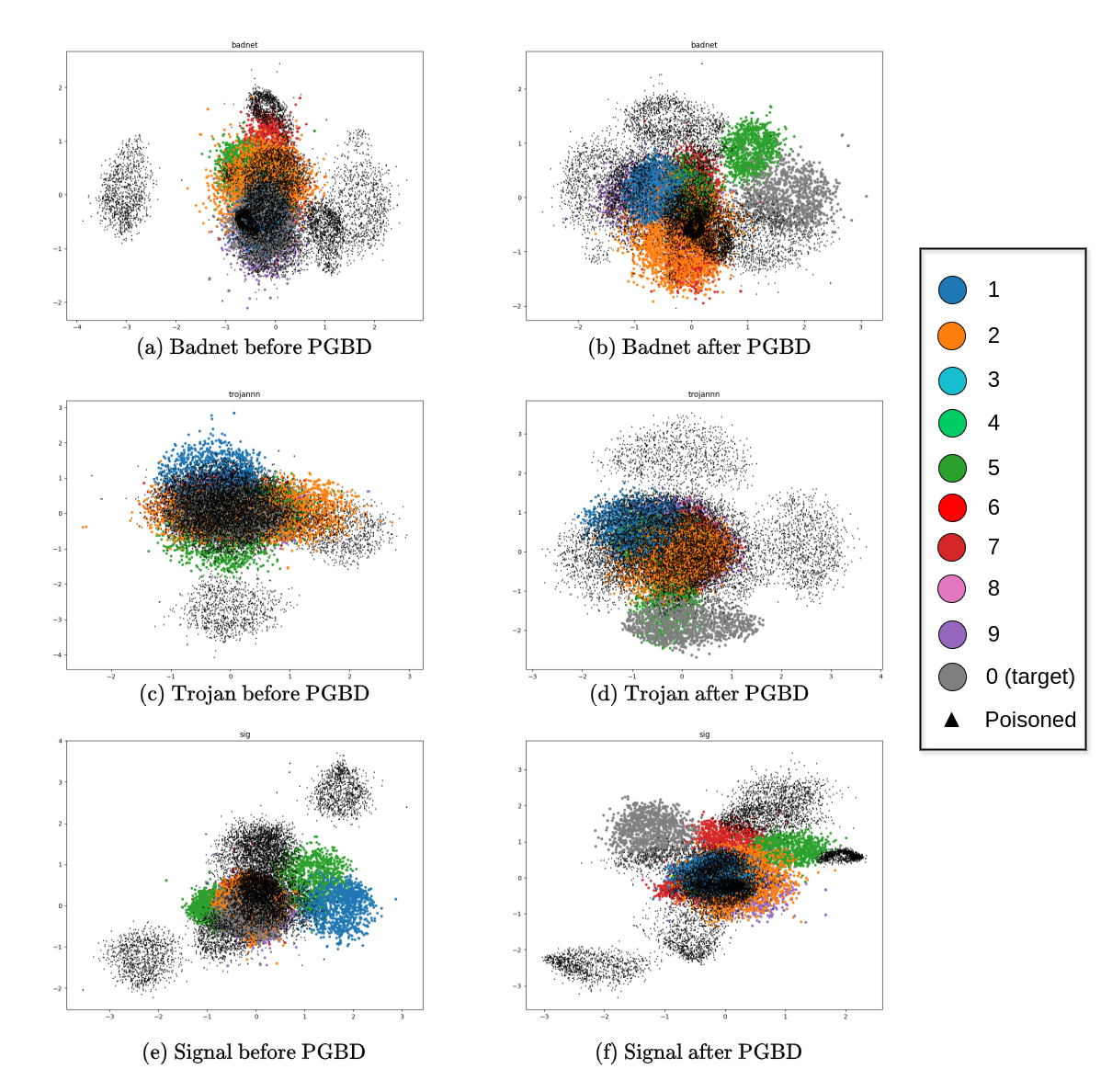}
    \caption{t-SNE visualizations of student activation space before and after PGBD defense for three attacks on CIFAR10.}
    \label{fig:tsne}
\end{figure*}

We show the t-SNE \cite{tsne} visualizations of the student model activations before and after PGBD in \Cref{fig:tsne}. Pre-defense plots show a common trend of poisoned samples clustering around the target class. Post-defense, we can clearly observe the dispersion of poisoned samples away from the target class and towards their respective classes. We also observe better class-wise clustering, which could be attributed to our method's prototype loss (\(L_p\)) usage.

\subsection{Spider plots}
\begin{figure}
    \centering
    \includegraphics[width=0.9\linewidth]{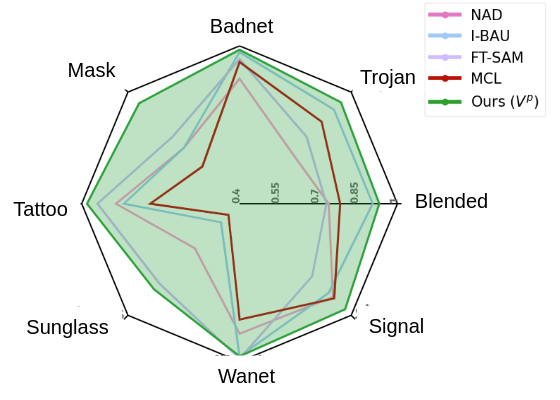}
    \caption{Spider plot of average DEM values over all datasets on 5 state of the art backdoor attacks. Our PGBD method beats previous works in all defense settings.}
    \label{fig:spider_dem_all}
\end{figure}

\begin{figure*}
    \centering
    \includegraphics[width=0.8\textwidth]{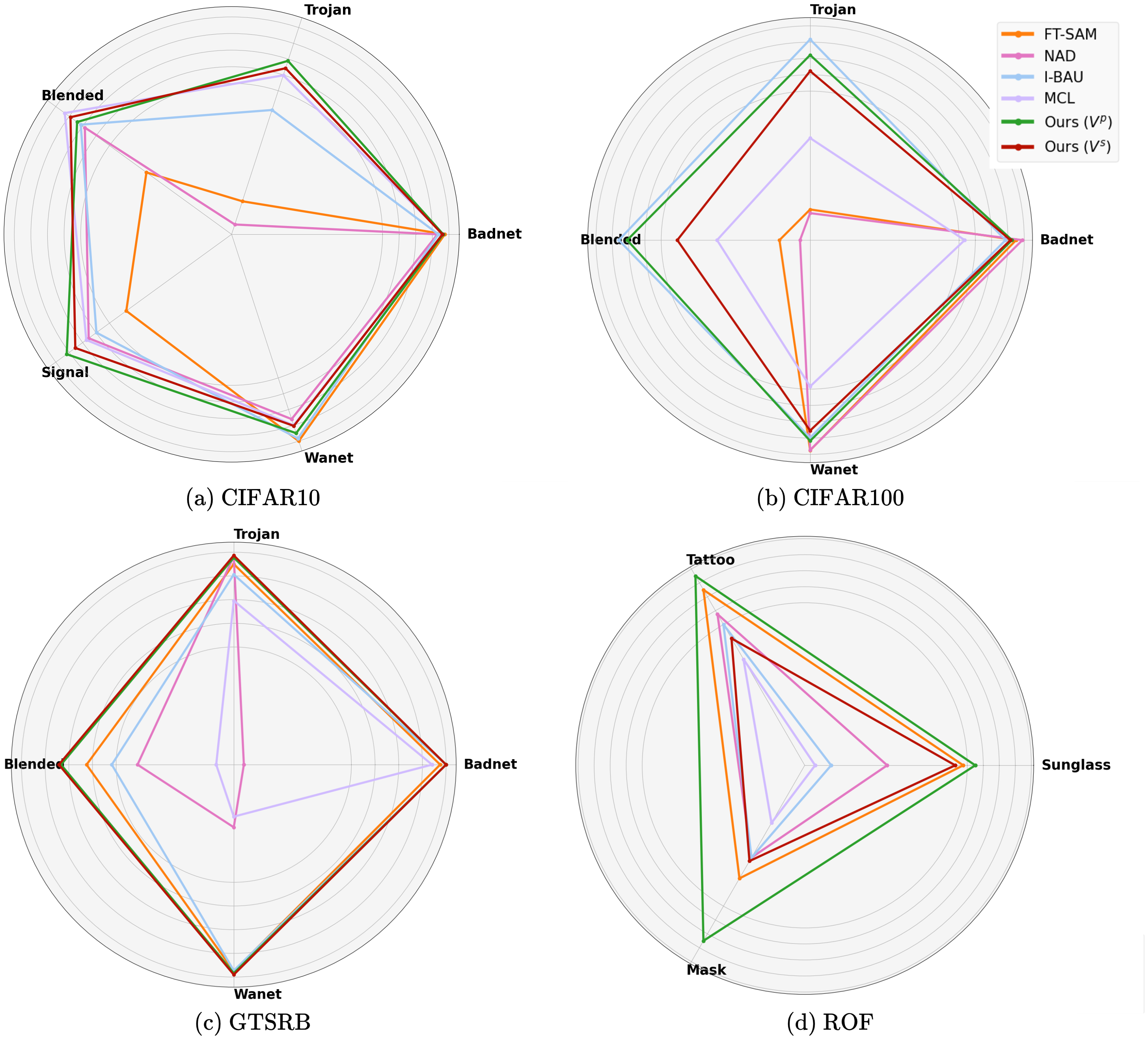}
  \caption{Spider plots of dataset wise DEM values. Refer to main paper teaser image for spider plot of overall mean results. More area inside the polygon represents better performance. We show outstanding performance on the unsolved semantic attacks (see (d)).}
  \label{fig:spider_plots}
\end{figure*}

\begin{figure*}
\centering
  \includegraphics[width=0.95\textwidth]{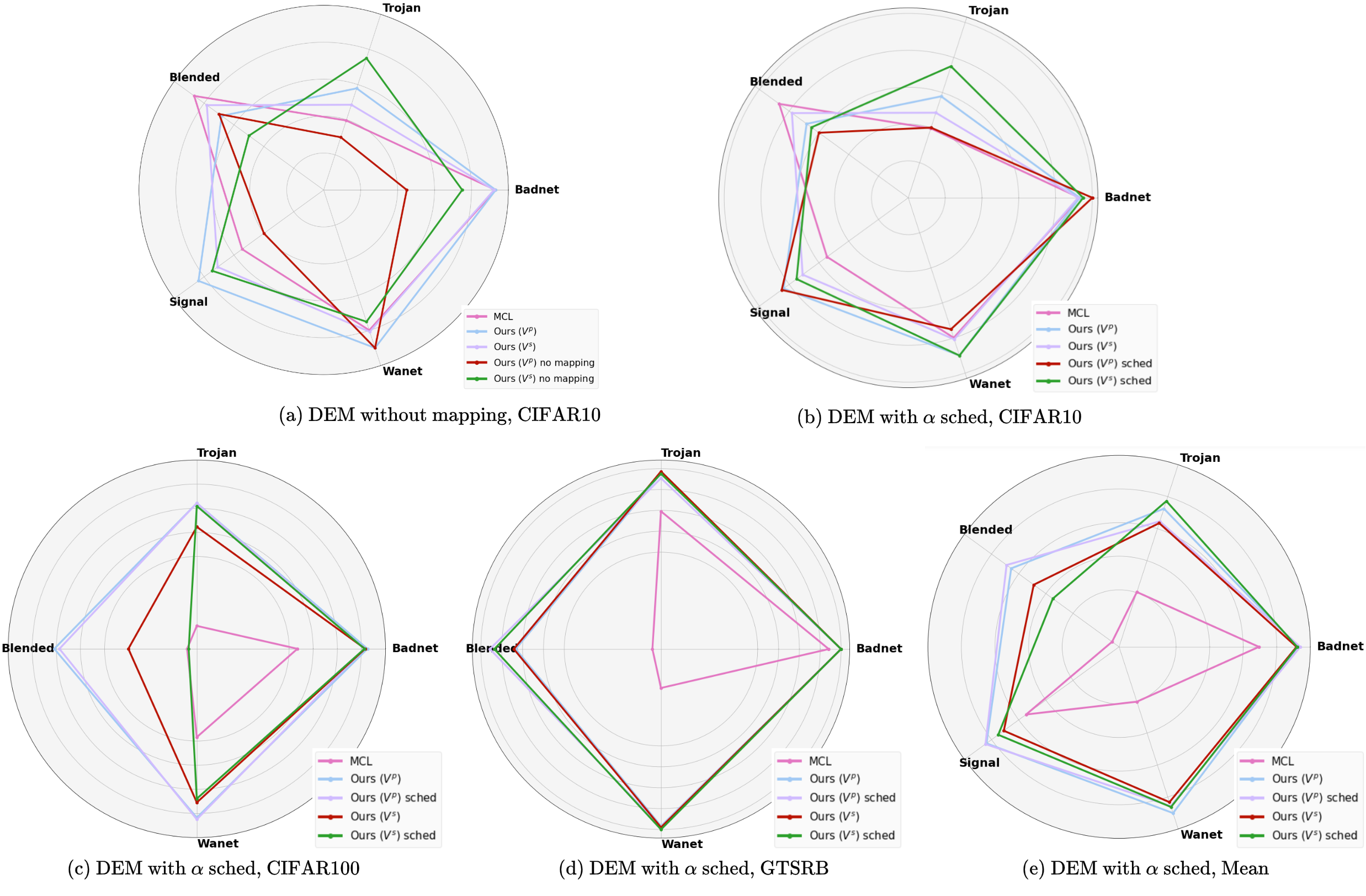}
  \caption{Spider plots of DEM values for ablations. Without large-model mapping results (top left) on CIFAR10, with \(\alpha\) scheduler results (top right) on CIFAR10, with \(\alpha\) scheduler results (bottom left) on CIFAR100,with \(\alpha\) scheduler results (bottom center) on GTSRB,with \(\alpha\) scheduler results mean across datasets (bottom right).}
  \label{fig:dem_all}
\end{figure*}

We visualize our results for easier comparison in the form of spider plots in \Cref{fig:spider_dem_all}. We additionally visualize dataset level results in \Cref{fig:spider_plots} and ablation results for both mapping and $\alpha$ scheduler in \Cref{fig:dem_all}. The spider plots show DEM values, therefore the larger the area enclosed, the better the overall performance across attacks. PGBD beats state of the art across attacks, additionally, we observe that hyperparameter tuning is helpful in some cases where our base method falls short.

\Paragraph{Recommendations}
Our method is dynamic as we update the prototypes after a certain interval. Considering this, our finetuning process still has scope for improvement, with the possibility of achieving a better global minima.We suggest using a early stopping condition in order to avoid over correction which impacts CA. One could use intermittent DEM or $\delta_a$ values as a metric to save best model checkpoints. Given that all our results are found within just 35 epochs, we would not recommend running for longer than that in search of a better output model. 

Given the flexibility of our method and hence multiple possible defense configurations, we would suggest the defender to choose hyperparameters according to their desired clean model (post-sanitization) behavior after analyzing the empirical results provided in the various tables.

\Paragraph{Limitations:}
The additional computational overhead of deriving and tracking the class prototypes while defending is a drawback of our method, particularly the NT-PGBD variant. This is expected given that PGBD is largely based on class-level manipulations of the activation space as opposed to previous sample-level defenses.

\Paragraph{Ethical Considerations:}
We deal directly with backdoor attacks which can raise several ethical questions. However, our intent is to {\em defend} against attacks, which is an ethically positive contribution. On the whole, research in attacks as well as defense brings more clarity and awareness to the community which should have a positive impact.

\section{Conclusion}
We additionally provide our semantic attack datasets and our code base as part of this supplementary. For the code base, we provide a backdoored model with the Badnet attack trained using code from \cite{wu2022backdoorbench} for easy startup. The implementation of DINOv1 feature extractor is taken from \cite{tschernezki22neural}. The code will be made public upon acceptance. Finally, we would like to thank the reviewers for their consideration and hope that the impact of our work has been presented thoroughly.

\end{document}